\pdfoutput=1
\documentclass[sigconf]{acmart}

\usepackage{booktabs} % For formal tables
\usepackage{tikz}
\usepackage{pgfplots}
\usepackage{textcomp}
\usepackage{siunitx}
\pgfplotsset{compat=1.12}
\setcopyright{rightsretained}
\copyrightyear{2018} 
\acmYear{2018} 
\setcopyright{acmcopyright}
\acmConference[ICVGIP 2018]{11th Indian Conference on Computer Vision, Graphics and Image Processing}{December 18--22, 2018}{Hyderabad, India}
\acmBooktitle{11th Indian Conference on Computer Vision, Graphics and Image Processing (ICVGIP 2018), December 18--22, 2018, Hyderabad, India}
\acmPrice{15.00}
\acmDOI{10.1145/3293353.3293396}
\acmISBN{978-1-4503-6615-1/18/12}

\begin{document}
\title{Cluster Loss for Person Re-Identification}
%\titlenote{Produces the permission block, and copyright information}
%\subtitle{Subtitle if Any}
%\subtitlenote{The full version of the author's guide is available as \texttt{acmart.pdf} document}

\author{Doney Alex}
%\authornote{Dr.~Trovato insisted his name be first.}
\orcid{https://orcid.org/0000-0002-7848-5461}
\affiliation{%
  \institution{Capillary Technologies}
  \streetaddress{P.O. Box 560068}
  %\city{Bangalore}
  %\state{Karnataka}
  \postcode{560068}
}
\email{doney.alex@capillarytech.com}

\author{Zishan Sami}
%\authornote{The secretary disavows any knowledge of this author's actions.}
\affiliation{%
  \institution{Indian Institute of Technology Kharagpur}
  %\streetaddress{P.O. Box 1212}
}
\email{zishansami102@iitkgp.ac.in}

\author{Sumandeep Banerjee}
%\authornote{Dr.~Trovato insisted his name be first.}
\affiliation{%
  \institution{Capillary Technologies}
  \streetaddress{P.O. Box 560068}
%   \city{Bangalore}
%   \state{Karnataka}
  \postcode{560068}
}
\email{sumandeep.banerjee@capillarytech.com}

\author{Subrat Panda}
%\authornote{Dr.~Trovato insisted his name be first.}
\affiliation{%
  \institution{Capillary Technologies}
  \streetaddress{P.O. Box 560068}
%   \city{Bangalore}
%   \state{Karnataka}
  \postcode{560068}
}
\email{subrat.panda@capillarytech.com}

% Add more authors as aboveif needed

% If the list of authors is too long for headers, add the following line:
\renewcommand{\shortauthors}{Alex et al.}
\settopmatter{printacmref=false} % Removes citation information below abstract
\pagestyle{plain} % removes running headers

\begin{abstract}
Person re-identification (ReID) is an important problem in computer vision, especially for video surveillance applications. The problem focuses on identifying people across different cameras or across different frames of same camera. The main challenge lies in identifying similarity of the same person against large appearance and structure variations, while differentiating between individuals. Recently, deep learning networks with triplet loss has become a common framework for person ReID. However, triplet loss focuses on obtaining correct orders on the training set. We demonstrate that it performs inferior in a clustering task. In this paper, we design a cluster loss, which can lead to the model output with a larger inter-class variation and a smaller intra-class variation compared to the triplet loss. As a result, our model has a better generalisation ability and can achieve a higher accuracy on the test set especially for a clustering task. We also introduce a batch hard training mechanism for improving the results and faster convergence of training.
\end{abstract}

\begin{CCSXML}
<ccs2012>
<concept>
<concept_id>10010147.10010178.10010224.10010245.10010255</concept_id>
<concept_desc>Computing methodologies~Matching</concept_desc>
<concept_significance>500</concept_significance>
</concept>
<concept>
<concept_id>10010147.10010178.10010224.10010225.10003479</concept_id>
<concept_desc>Computing methodologies~Biometrics</concept_desc>
<concept_significance>300</concept_significance>
</concept>
<concept>
<concept_id>10010147.10010257.10010293.10010294</concept_id>
<concept_desc>Computing methodologies~Neural networks</concept_desc>
<concept_significance>300</concept_significance>
</concept>
</ccs2012>
\end{CCSXML}

\ccsdesc[500]{Computing methodologies~Matching}
\ccsdesc[300]{Computing methodologies~Biometrics}
\ccsdesc[300]{Computing methodologies~Neural networks}
%\keywords{ACM proceedings, \LaTeX, text tagging}

\maketitle

\def\eg{\emph{e.g}}
\def\Eg{\emph{E.g}}
\def\etal{\emph{et al. }}

\section{Introduction}

Person re-identification (ReID) is an important problem in computer vision especially for video surveillance applications. Major challenges include variations of lighting conditions, poses, viewpoints, blurring effects, image resolutions, camera settings, occlusions, background etc. The person ReID task is similar to image retrieval or face recognition in many ways. With advancements in deep learning, significant improvements have been made in the areas of image retrieval. There are many works in person ReID which were motivated from face recognition. One such example on which many person ReID methods are based on is FaceNet\cite{facenet}, a convolutional neural network (CNN) used to learn an embedding for faces. The key component of FaceNet is to use the triplet loss, as introduced by Weinberger and Saul\cite{tripletMotivation}, for training the CNN as an embedding function. The triplet loss optimizes the embedding space such that data points with the same identity are closer to each other than those with different identities.
\begin{figure}
    \includegraphics[width=.8\linewidth,height=1.2\linewidth]{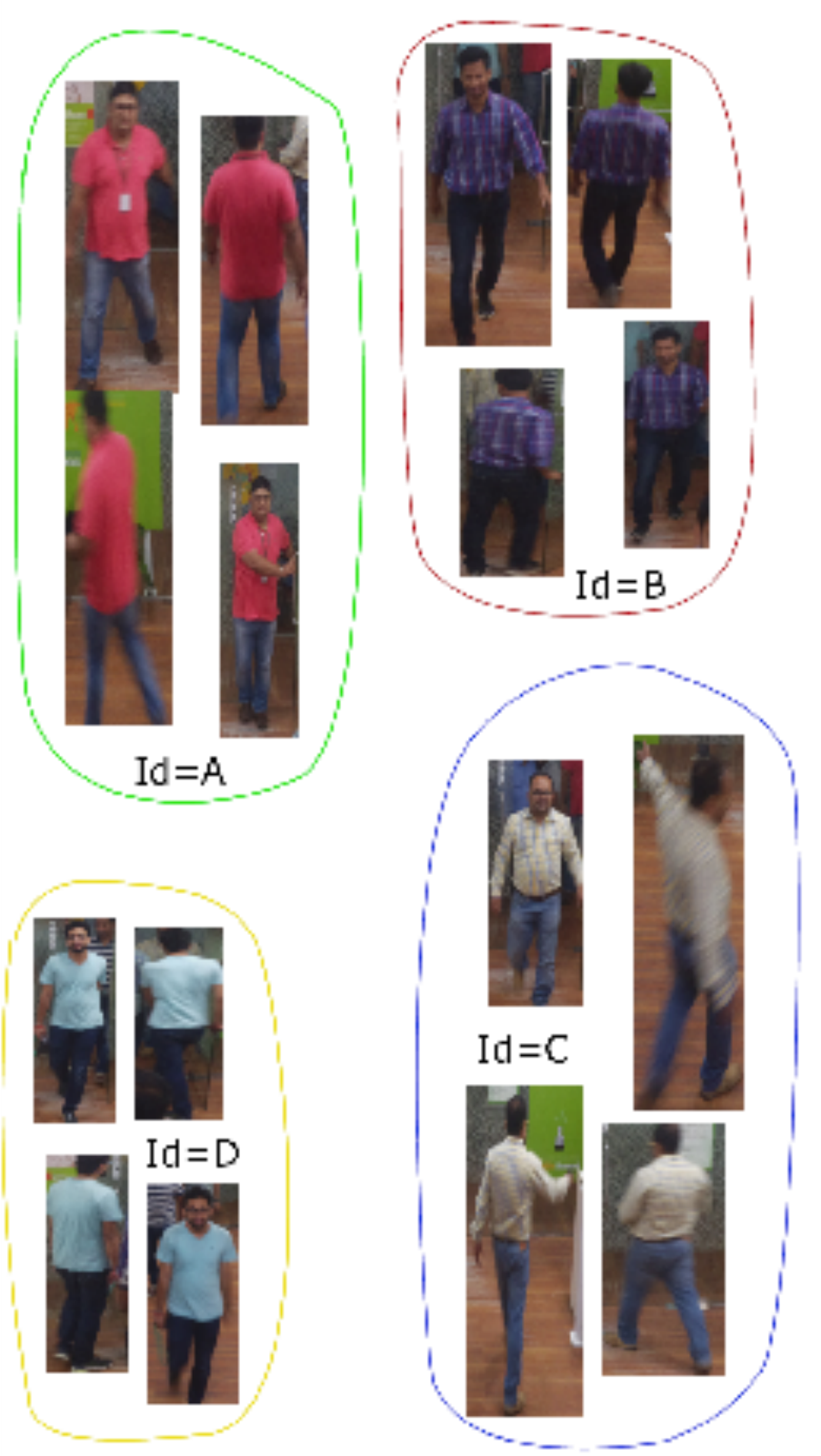}
    \caption{Clustering Different People Images. }\label{fig:clustImages}
\end{figure}
Even though there are a variety of approaches in loss functions such as classification loss, with a combination of verification loss in some cases \cite{triplet_modi9,classification_verfic_2,classification_verfic_3,classification_verfic_4} or other losses like DeepLDA \cite{deepLDA}, triplet losse and its variations\cite{triplet_modi1,triplet_modi2,triplet_modi3,triplet_modi4,triplet_modi5,triplet_modi6,triplet_modi7,triplet_modi8,triplet_modi9,triplet_modi10,TriNet,quadreplet_loss} seem to be the most common and successful approach.  Cumulative Matching Characteristic curve   which follows rank-n criteria  is  the most common \cite{cumMAtcCurve2,cumMAtcCurve1,survey1} method used for performance evaluation of person ReID. Recent deep learning approaches \cite{ranking,triplet_modi4,triplet_modi7,triplet_modi5,triplet_modi2,TriNet} usually treat person ReID as a ranking task and apply a triplet loss to address the problem. The main purpose of the triplet loss, which is motivated in the context of nearest-neighbour classification\cite{facenet,tripletMotivation}, is to obtain a correct order for each probe image and distinguish identities in the projected space. But these methods seem to perform inferior in clustering tasks. The underlying reason is that the model trained by a triplet loss would still cause a relatively large intra-class variation\cite{triplet_modi4,triplet_bad_intra}.

In this paper we introduce cluster loss, motivated by Linear Discriminant Analysis and K-Means clustering. While triplet loss tries to minimize the distance between similar images, our clustering loss tries to minimize the distance between images to the mean of their class and maximize the distance between the means of other classes. This results in all images of same identity to come together to form a cluster and the clusters to stay separated. Hence our model is capable of achieving a smaller intra-class variation and a larger inter-class variation with significant performance on the test set. 

Many recent deep learning approaches treat the person ReID as a ranking task and use rank-n criteria for performance evaluation. Clustering is also an important application of person ReID. For eg, in a scenario where continuous feed from a surveillance camera captures a moving person in multiple frames, images belonging to similar identity would need to be grouped ( as shown in Fig \ref{fig:clustImages}), in order to do any analysis/recognition. Therefore we wanted to evaluate the output of person ReID network using clustering algorithms. We used a simple sequential  clustering (explained in section \ref{seqeval}) to evaluate the performance in a clustering scenario. It was observed that our method outperforms the existing methods for Person ReID in a clustering task by a huge margin.

%%%%%%%%%%%%%%%%%%%%%%%%%%%%%%%%%%%%%%%%%
\section{Related Work}
Most developments in person ReID problem concentrate on feature extraction and similarity measurement. Traditional feature extraction techniques largely use colour histograms, local binary patterns, texture filters etc. Gray and Tao \cite{colorHist1} use 8 colour channels (RGB, HS, and YCbCr)
and 21 texture filters on the luminance channel, and the
pedestrian is partitioned into horizontal stripes. A number
of later works \cite{colorHist2,colorHist3,colorHist4}employ the same set of features as \cite{colorHist1}. Similarly, Mignon et al. \cite{colorHist5} built the feature vector
from RGB, YUV and HSV channels and the LBP texture
histograms in horizontal stripes. Most hand crafted features rely on colour histograms and texture filters but there are works which use complex features like SIFT\cite{sift} or local maximal occurrence (LOMO) descriptor\cite{lomo}, which includes the colour and SILTP histograms. Another  choice is the attribute-based features which  are more robust to image translations compared to low-level descriptors. The low-level  features like colour,texture or category labels are used to train the attribute classifiers\cite{attr1,attr2}.

In a ReID system with hand crafted features, a good distance metric is critical for its success,  because the high-dimensional visual features may not capture the    invariant factors under sample variances. In person ReID many works fall into the scope of supervised global distance metric learning. The task of global metric learning is to keep all the vectors of the same class closer while pushing vectors of different classes further apart. The most commonly used formulation is based on the class of Mahalanobis distance functions and its modifications~\cite{metric1,metric2}, which generalizes Euclidean distance using linear scalings and rotations of the feature space. One popular metric learning method is  KISSME~\cite{kissme}  which is based on Mahalanobis distance and the decision on whether a pair is similar or not is formulated as a likelihood ratio test. Apart from the methods that use Mahalanobis distance, some use other learning tools such as support vector machine (SVM) or boosting. In ~\cite{svm1}, a structural SVM is employed to combine different colour descriptors at decision level and in ~\cite{svm2}, a specific SVM is learned for each training identity and map each testing image to a weight vector inferred from its visual features. Gray \etal \ propose using the AdaBoost algorithm to select and combine many different kinds of simple features into a single similarity function in ~\cite{colorHist1}.

In traditional methods, feature extraction and similarity measurement are treated independently, because of which those methods could not reach the performance level of CNN based systems, where the end-to-end system can be globally optimized via back-propagation. The major bottleneck of deep learning methods in ReID was the lack of training data. With the advancement of deep learning in almost all fields and the increasing availability of datasets, CNN based methods which automatically learn features and metrics became common in ReID and hence the handcrafted features and metrics struggle to keep top performance widely, especially on large scale datasets.

Most CNN-based ReID methods focus on the Siamese model. In ~\cite{siam1}, an input image is partitioned into three overlapping horizontal parts, and the parts go through two convolutional layers and a fully connected layer which fuses them and outputs a vector for the image and the similarity of the two output vectors are computed using the cosine distance. There are many modified versions of Siamese model like ~\cite{siam2} in which cross- input neighbourhood difference features are computed, which compares the features from one input image to features in neighbouring locations of the other image or like ~\cite{siam3} which uses product to compute patch similarity in similar latitude. Meanwhile, there are methods ~\cite{siam3,siam2,class1} which tackle the person ReID problem using a classification/identification mode, which makes full use of the re-ID labels. In ~\cite{classficx}, training identities from multiple datasets jointly form the training set and a softmax loss is employed in the classification network. Some of them use a softmax layer with the cross-entropy loss in their networks ~\cite{siam3,class1}. The cross-entropy loss can well represent the probability that the two images in the pair are of the same person or not. Some other methods use a margin-based loss~\cite{triplet_modi5}, which builds a margin to maintain the largest separation between positive and negative pairs. For instance, Varior \etal ~\cite{lstm} incorporate long short-term memory (LSTM) modules into a Siamese network. LSTMs process image parts sequentially so that the spatial connections can be memorized to enhance the discriminative ability of the deep features.

While Siamese networks based works use image pairs, Cheng \etal ~\cite{triplet_modi4} design a triplet loss function that takes three images as input. A drawback of the Siamese model with triplet loss is that it does not make full use of ReID annotations. These models only needs to consider pairwise (or triplet) labels. Telling whether an image pair is similar (belong to the same identity) or not is a weak label in ReID. Sometimes triplet loss based networks may produce disappointing results especially when applied naively. An essential part of learning using the triplet loss is the mining of hard triplets, as otherwise training will quickly stagnate. However, mining such hard triplets is time consuming and it is unclear what defines "good" hard triplets ~\cite{triplet_modi6}. Even worse, selecting too many hard triplets too often makes the training unstable.
Another major caveat of the triplet loss is that as the dataset gets larger, the possible number of triplets grows cubically, rendering a long enough training impractical. As training progress, the transformation output relatively quickly learns to correctly map most trivial triplets, rendering a large fraction of all triplets uninformative. Hence we introduce cluster loss, which tries to minimise not just the distance between the similar pairs but the distance between all similar images with respect to their mean and increases the distance between the means thereby making sure each each unique cluster stay apart.

\section{The proposed approach}
We strive for an embedding $f(x)$, from an image
$x$ into a feature space $R^d$, a $d$-dimensional Euclidean space, such that the squared distance between all person images, independent of imaging conditions, belonging to the same identity is small to form a cluster and the squared distance between clusters is large. The triplet loss is motivated in the context of nearest-neighbour classification. We introduce our clustering loss taking motivation from K-Means clustering and Linear Discriminant Analysis.
\subsection{Network Architecture}
\begin{figure}
    \includegraphics[width=\linewidth]{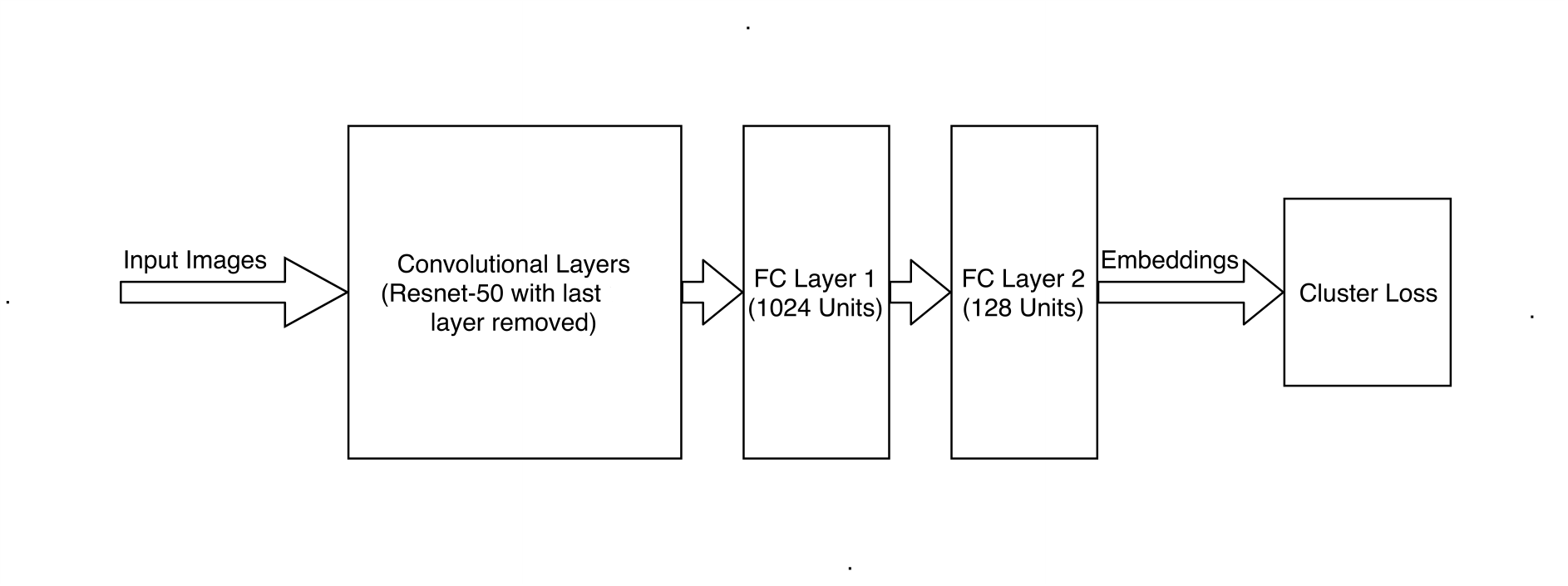}
    \caption{The proposed Architecture. }\label{fig:arch}
\end{figure}
We use the ResNet-50 architecture for the convolutional layers similar to that used in ~\cite{TriNet}. We experimented with other networks like VGG~\cite{vgg} and GoogLeNet~\cite{incpetion} but the results were similar to that of ResNet-50. The ResNet-50 was chosen because it is computationally less demanding compared to other deeper networks like VGG and GoogLeNet. In ResNet-50, the last layer is discarded and we add two fully connected layers for our task as shown in Fig \ref{fig:arch}. The first has 1024 units, followed by batch normalization~\cite{batchnorm} and ReLU~\cite{relu}, the second goes down to 128 units, our final embedding dimension. The network had about 25.74 M parameters. The batch size is limited to 256 containing P = 16 persons with K = 16 images each. We chose learning rate $\epsilon_0 = 3$x$10^{-5}$ with learning rate decay starting after 25000 iteration for a total of 50000 iterations and Adam optimizer~\cite{adam} with the default hyper-parameter values ($\epsilon = \epsilon_0 , \beta1 = 0.9, \beta2 = 0.999$) for the experiments. We performed all our experiments using the Tensorflow\cite{tesnsorflow} framework.
\subsection{Loss Function}
We use Euclidean distance as metric for separation between two samples in the transformed space $R^d$. Triplet loss is used for performance comparisons. Hence we are going to introduce triplet loss first.
\subsubsection{Triplet Loss}
In triplet loss we create a collection of triplets such that we select an anchor image $x_i^a$, a positive image $x_i^p$ which is another image of same person and a negative image $x_i^n$ of a different person. The triplet loss wants to keep  $x_i^a$ and $x_i^p$ closer.  
For every set $i$, we want 
\begin{equation}
||f(x_i^a) -f(x_i^p)||^2_2 + \alpha < ||f(x_i^a) -f(x_i^n)||^2_2\label{equation:trip1 } 
\end{equation}

Hence the loss that is being minimized is 
\begin{equation}
L_{trp} =\sum_i^N [||f(x_i^a) -f(x_i^p)||^2_2  - ||f(x_i^a) -f(x_i^n)||^2_2+ \alpha]\label{equation:trip2 } 
\end{equation}
In Eq  \ref{equation:trip1 }, the triplet loss adopts the Euclidean distance to measure the similarity of extracted features from two images. The major challenge with  triplet loss is that as the dataset gets larger, the possible number of triplets grows cubically, rendering a long enough training impractical. The transformation function $f$  relatively quickly learns to correctly map most trivial triplets, rendering a large fraction of all triplets uninformative.

\subsubsection{Cluster Loss}
We take motivation from K-Means clustering and Linear Discriminant Analysis. The target is to minimise intra class variations and to maximize the inter class variations. In a batch of $N$ images with $P$ person identities containing $K$ images of each person, for a person identity $i\in P$, mean $f^m_i$ in feature space $R^d$ is,
\begin{equation}
 f^m_i = \frac{\sum^K f(x)} {K}
 \label{equation:clumean } 
\end{equation}

Intra class variation for an identity is represented by the distance of each sample of that identity to the mean of that identity. Hence for an identity $i$, intra class variation $d^{intra}_i$ is given by
\begin{equation}
 d^{intra}_i = {\sum_k ||f(x) -f^m_i||^2_2 } 
 \label{equation:cluintra } 
\end{equation}

Similarly, inter class variation for an identity is represented by the distance of the mean of that identity to means of all other identities. Hence for an identity $i$, inter class variation $d^{inter}_i$ is given by
\begin{equation}
 d^{inter}_i = {\sum_{\forall  i_d \in P,i_d\ne i } ||f^m_i -f^m_{i_d}||^2_2 } 
 \label{equation:cluiner } 
\end{equation}

The task is to minimise intra class distances and maximise the inter class distances. Hence the loss that is being minimised is 
\begin{equation}
L_c = \frac{\beta\sum_i^P d^{intra}_i }{\gamma + \sum_i^P d^{inter}_i }
\label{equation:clusLoss } 
\end{equation}

The summation term in the numerator in Eq. \ref{equation:clusLoss } accumulates PK distances where as summation  in the denominator accumulates P(P-1) distances. Hence $\beta$ is a hyper parameter which act as a normalising constant and $\gamma$ is a very small value.

\subsection{Batch Hard Training}
The loss function shown by Eq.\ref{equation:clusLoss } 
describes the basic concept of cluster loss. We strive to minimise the intra class distance which is measured as distance of samples of a class with respect to their mean, at the same time maximising inter class distances which is measured as distances between the means of different classes. Although Eq. \ref{equation:clusLoss } is a good representation of cluster loss, when we trained the network with that particular loss function, the results were not promising and the number of iterations required for convergence was very high. This is because the loss contained equal contributions from all samples. This is similar to training using triplet loss without mining hard triplets. The transformation $f$ relatively quickly learns to correctly map most trivial samples, rendering a large fraction of all samples uninformative. Thus mining hard positive/negative samples becomes crucial for learning. Intuitively, being told over and over again that people with differently coloured clothes are different persons does not teach one anything, whereas seeing similarly looking but different people (hard negatives), or pictures of the same person in wildly different poses or from different camera angles (hard positives) dramatically helps in understanding the concept of re-identification.

So we modified the loss function in such a way that it does not take cumulative contribution from all images in a batch but the samples which contribute most to the loss, so that the correction step by minimization affects those samples which have the maximum error. Hence only hard samples contribute directly to the loss function. In this approach for the new $d^{intra}_{i}$ of an identity $i$, we take the sample which lies farthest from the mean $f^m_i$ and take the corresponding distance as $d^{intra}_{i}$.
\begin{equation}
 d^{intra}_{i} = {\max_K ||f(x) -f^m_i||^2_2 } 
 \label{equation:cluintraBh1 } 
\end{equation}
For the new $d^{inter}_{i}$ for an identity $i$, we take the distance with that mean which is closest to the mean of considered identity.
\begin{equation}
 d^{inter}_i = {\min_{\forall  i_d \in P,i_d\ne i } ||f^m_i -f^m_{i_d}||^2_2 } 
 \label{equation:cluinetrBh1 } 
\end{equation}
The final loss function to be minimised is 
\begin{equation}
Lb_c = \sum_i^P \max((d^{intra}_{i}  -d^{inter}_{i} +\alpha),0)
\end{equation}

Mining hard samples ensures that the training converges fast and better results. In ~\cite{TriNet} a method for mining hard triplets is described which gives better results compared to other triplet loss based methods. The downside of basing the loss function only on few triplets is that, the transformation function is adjusted based only on the distance between those samples. In our method, even though we consider only hard samples, since their distances are calculated with respect to mean, every sample contributes indirectly. Hence with every iteration, the transformation adjusts to decrease the distance within the clusters while making sure that the clusters stay far apart.
\section{Experiments and Results}
We focused on two types of performance evaluations. 1) Performance for a ranking task and 2) Performance for sequential clustering task. The datasets we employed were Market-1501~\cite{market}, one of the  largest person ReID datasets currently available and CUHK03~\cite{siam3} dataset. The Market-1501 dataset contains bounding boxes from a person detector which have been selected based on their intersection-over-union overlap with manually annotated bounding boxes. It contains 32668 images of 1501 persons, split into train/test sets of 12936/19732 images as defined by~\cite{market}. We also show results on the CUHK03~\cite{siam3} dataset which contain 13164 images of  1360 identities.
\begin{center}
\begin{figure}
 \begin{tabular}{|p{.44\linewidth}|p{.12\linewidth}|p{.12\linewidth}|p{.12\linewidth}|}
 \hline
  &mAP &rank-1 &rank-5 \\
 \hline
  TriNet~\cite{TriNet}&69.14 &84.92& 94.21 \\ 
  \hline
  LuNet~\cite{TriNet}&60.71&81.38&92.34\\ 
  \hline
  CAN~\cite{triplet_modi10}&35.9&60.3&-\\ 
  \hline
 IDE (R) + ML~\cite{rerank}&49.05&63.60&-\\
  \hline
  LOMO + Null Space~\cite{table1}&29.87&55.43&-\\ 
  \hline
  APR (R, 751)~\cite{table2}&64.67&84.29&93.20\\
  \hline
  JLML~\cite{classification_verfic_4}&65.5&85.1&-\\
  \hline
 Latent Parts (Fusion)~\cite{table3}&57.53&80.31&-\\
  \hline
  Gated siamese CNN~\cite{lstm}&39.55&65.88&-\\ 
  \hline
  DTL~\cite{classification_verfic_2}&65.5&83.7&-\\
  \hline
 \textbf{Our Method}&\textbf{71.54}&\textbf{86.07}&\textbf{94.98}\\
  \hline
 \textbf{Our Method (Re-ranked)}&\textbf{82.84}&\textbf{87.97}&\textbf{94.18}\\
  
 \hline
\end{tabular}
  \caption{Performance of latest methods for ReID on Market-1501 SQ dataset.}
  \label{fig:marKetRank}
\end{figure}
\end{center}
Augmenting training data is a common practice. We performed random crops and random horizontal flips during training. Similar to the augmentation steps in TriNet~\cite{TriNet}, we resize all images of size H x W to 1$\frac{1}{8}$(H x W),
of which we take random crops of size H x W , keeping their aspect ratio intact. We set H = 256, W = 128 on Market-1501  and H = 256, W = 96 on CUHK03. We apply test-time augmentation in  our experiments. From each image, we deterministically create five crops of size H x W : four corner crops and one center crop, as well as a horizontally flipped copy of each. The embeddings of all these ten images are then averaged, resulting in the final embedding for a person. We also experimented with transfer learning. We initialized our network with weights of existing network which was trained for ReID task like TriNet~\cite{TriNet} and this yielded a better result which converged with fewer iterations.

\subsection{Performance for ranking task}
We evaluated the performance for ranking task on both Market-1501~\cite{market} and CUHK03~\cite{siam3} datasets. We used the standard evaluation, namely the mean average precision score (mAP) and the cumulative matching curve (CMC) at rank-1 and rank-5. We followed the evaluation codes provided by ~\cite{rerank} and ~\cite{TriNet}.
\begin{center}
\begin{figure}
 \begin{tabular}{|p{.49\linewidth}|p{.18\linewidth}|p{.18\linewidth}|}
 \hline
  &rank-1 &rank-5\\
 \hline
  TriNet~\cite{TriNet}&87.58&98.17 \\ 
  \hline
  kLFDA~\cite{chuckTable}&48.20&59.34\\ 
  \hline
 GatedSiamese~\cite{lstm}&68.10&88.10\\
  \hline
 GOG~\cite{chuckTable2}&67.30&91.00\\
  \hline
 DGD~\cite{classficx}&80.50&94.90\\
 \hline
 Ensembles~\cite{triplet_modi3}&62.10&89.10\\
  \hline
 DeepLDA~\cite{deepLDA}&63.23&89.95\\
  \hline
 NullReid~\cite{table1}&58.90&85.60\\
  \hline
 IDLA~\cite{siam2}&54.74&86.50\\
  \hline
 Quadruplet~\cite{quadreplet_loss}&75.53&95.15\\
 \hline
 \textbf{Our Method}&\textbf{87.87}&\textbf{98.79}\\
 \hline
\end{tabular}
  \caption{Performance of latest methods for ReID on CUHK03 dataset.}
  \label{fig:ChuckRank}
\end{figure}
\end{center}
Table ~\ref{fig:marKetRank} compares our results to a set of related, top performing approaches on Market-1501 with single query. We also evaluated how our model performs when combined with the re-ranking approach by Zhong \etal ~\cite{rerank}. This can be applied on top of any ranking methods and uses information from nearest neighbours in the gallery to improve the ranking result.
Table ~\ref{fig:ChuckRank} compares our results to a set of related, top performing approaches on CUHK03. It is evident that all deep learning based methods outperform the traditional methods by a huge margin. The TriNet~\cite{TriNet} which is based on triplet loss with an improvement in hard mining of samples seems to perform best among all existing methods. Our method performs slightly better than TriNet in the ranking task.
\begin{figure}
    \includegraphics[width=\linewidth,,height=.8\linewidth]{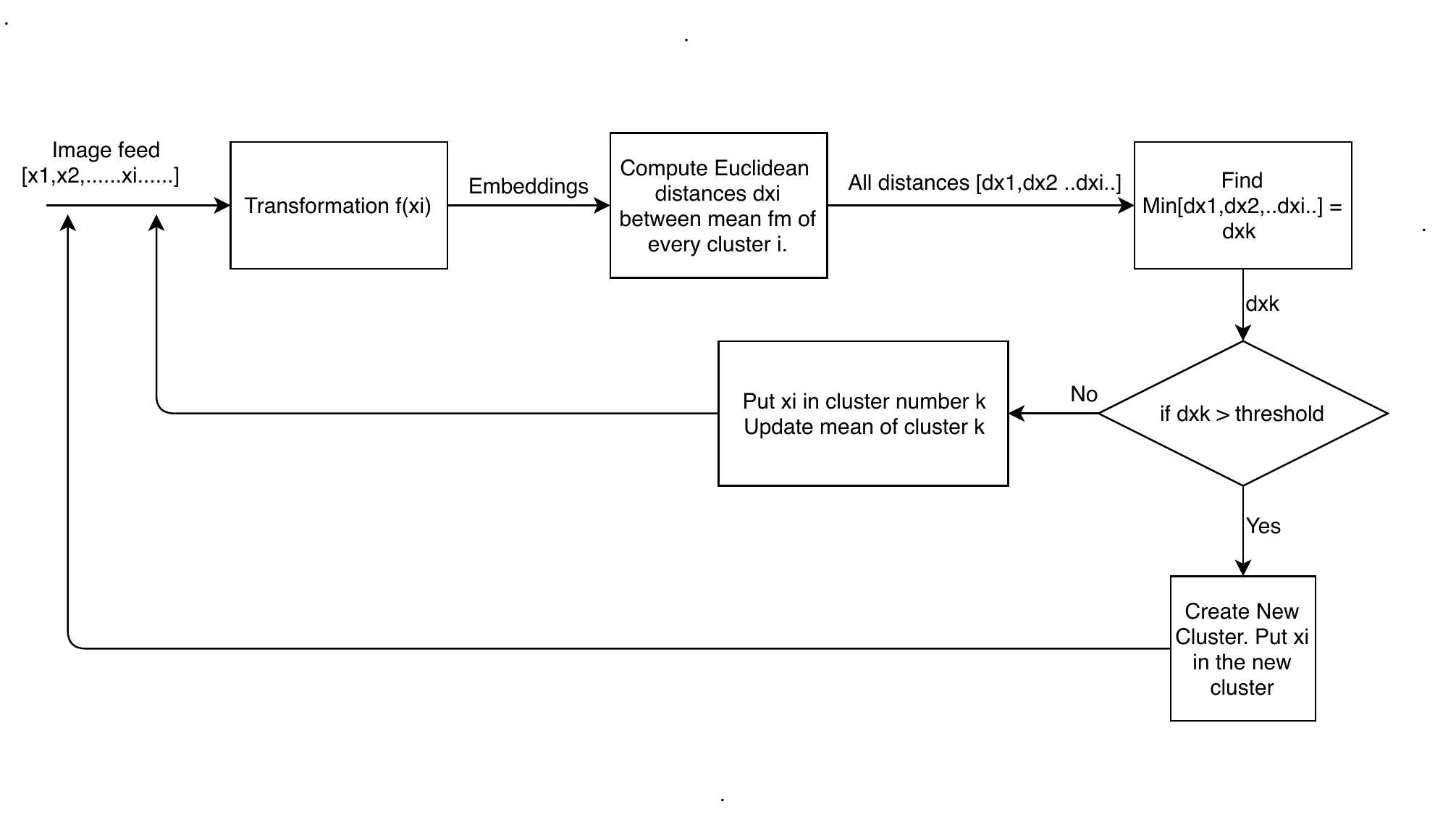}
    \caption{Sequence clustering process. }\label{fig:seqclust}
\end{figure}

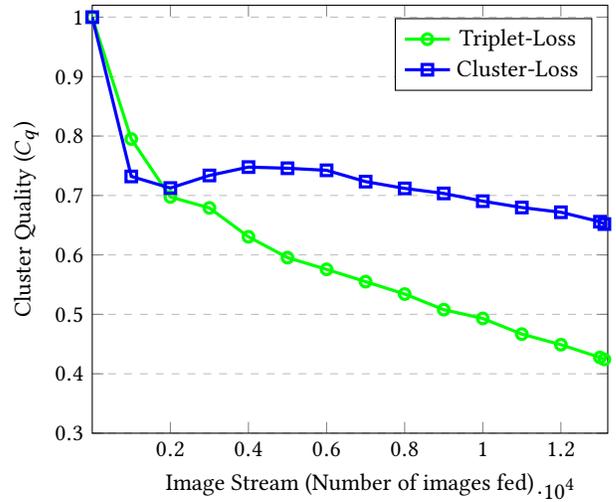
\begin{figure}
\begin{tikzpicture}
\begin{axis}[
    xlabel={Image Stream (Number of images fed)},
    ylabel={Cluster Quality ($C_q$)},
    xmin=0, xmax=13200,
    ymin=0.3, ymax=1.02,
    %point to be marked on axis
    xtick={2000,4000,6000,8000,10000,12000,14000},
    ytick={0.3,0.4,0.5,0.6,0.7,0.8,0.9,1},
    legend pos=north east,
    ymajorgrids=true,
    grid style=dashed,
]
 
 \addplot[
    color=green,
    mark=o,
    very thick
    ]
    %points which will be marked(x,y)
    coordinates {
    (0,1.0)(1000,0.795)(2000,0.6975)(3000,0.679)(4000,0.6305)(5000,0.5954)(6000,0.5756666666666667)(7000,0.555)(8000,0.534125)(9000,0.5078888888888888)(10000,0.4931)(11000,0.4667272727272727)(12000,0.4489166666666667)(13000,0.42746153846153845)(13113,0.424159231297186)
    };
    
\addplot[
    color=blue,
    mark=square,
    very thick
    ]
    coordinates {
    (0,1.0)(1000,0.732)(2000,0.7125)(3000,0.7336666666666667)(4000,0.74775)(5000,0.7458)(6000,0.7423333333333333)(7000,0.7232857142857143)(8000,0.711875)(9000,0.7034444444444444)(10000,0.6904)(11000,0.6797272727272727)(12000,0.67175)(13000,0.6558461538461539)(13113,0.6518721879051323)
    };
  \legend{Triplet-Loss, Cluster-Loss} 

\end{axis}

\end{tikzpicture}

    \caption{Variation of Cluster Quality ($C_q$) with the image feed }\label{fig:classClusterQuality}
\end{figure}
\subsection{Performance for sequential clustering task}
We wanted to evaluate the performance of our method for a clustering task. We did a simple sequential clustering explained in Fig \ref{fig:seqclust}, in which images ($x_i, x_2,....x_i....)$ are fed in sequence. For an image $x_i$, after passing through the network to find the transformation $f(x_i)$, Euclidean distances $d_i$ are computed with the means $f_m$ of
\begin{figure}
\begin{tikzpicture}
\begin{axis}[
    xlabel={Image Stream (Number of images fed)},
    ylabel={Rand Index ($R_{id}$)},
    xmin=0, xmax=13200,
    ymin=0.3, ymax=1.02,
    %point to be marked on axis
    xtick={2000,4000,6000,8000,10000,12000,14000},
    ytick={0.3,0.4,0.5,0.6,0.7,0.8,0.9,1},
    legend pos=north east,
    ymajorgrids=true,
    grid style=dashed,
]
 
 \addplot[
    color=green,
    mark=o,
    very thick
    ]
    %points which will be marked(x,y)
    coordinates {
    (0,1.0)(1000,0.7920066729127716)(2000,0.6693917997928762)(3000,0.6433901380513886)(4000,0.5810033099547114)(5000,0.5442987521530491)(6000,0.5153614026295482)(7000,0.49046729111870413)(8000,0.4588613473966029)(9000,0.4311305758240738)(10000,0.4085584144423138)(11000,0.38392412947880583)(12000,0.3620640683615314)(13000,0.3374110395729094)(13113,0.3331419558364587)
    };
    
\addplot[
    color=blue,
    mark=square,
    very thick
    ]
    coordinates {
    (0,1.0)(1000,0.7210134576787723)(2000,0.6864938942997069)(3000,0.7158985504649866)(4000,0.7161343534631847)(5000,0.7114509701454763)(6000,0.7076399792677122)(7000,0.6857055237156947)(8000,0.6637628838835034)(9000,0.6430834411929495)(10000,0.6236588238355382)(11000,0.6072225400090026)(12000,0.5931627215987935)(13000,0.5722698876353925)(13113,0.5629535057339076)
    };
  \legend{Triplet-Loss, Cluster-Loss} 
\end{axis}
\end{tikzpicture}
    \caption{Variation of Rand Index ($R_{id}$) with the image feed }\label{fig:classClusterRIndeQuality}
\end{figure}
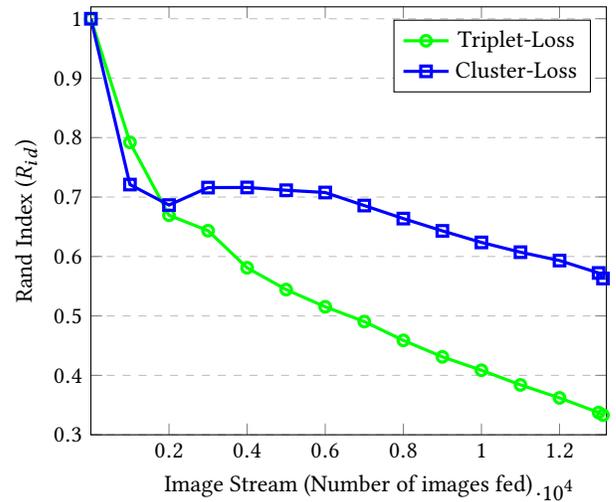all existing classes and the minimum among them is taken as $d_k$ ( as the minimum distance was with class $k$). If $d_k$ is less than threshold $th$, image $x_i$ is marked as belonging to class $k$ and mean of class $k$ is updated with $f(x_i)$. If $d_k$ is greater than or equal to $th$, $x_i$ is marked as a new class, increasing the total number of classes by 1. We used the test set of Market-1501~\cite{market} data set for the experiments after removing the "distractor" and "junk" images. To prepare the feed, we randomly select 4 to 6 identities (from a total of 750 identities) from the set and then shuffle all of their images and then push it to the feed. This is done to resemble the real life cases where multiple people are passing in front of a camera. Triplet loss based TriNet~\cite{TriNet} gave highest accuracy for ranking task among all previously existing methods. So we compared the sequential clustering performance of our method with TriNet by replacing the transformation network $f(x)$  with that of Trinet and created embeddings. We used two metrics for measuring the clustering accuracy.
1) Cluster Quality ($C_q$): At any stage of the image feed, we know the true identity of every image that has been fed to the clustering algorithm until that point. So we try to map identities to all existing clusters. The criterion for tagging a cluster with an identity is that, maximum number of images in that cluster belong to that particular identity ie an identity $I$ is assigned to a particular cluster $C$ if the maximum number of images in that cluster belong to $I$. Same identity can be assigned to more than one cluster. In such cases we take the cluster with most number of images belonging to that identity, and mark the other clusters as unassigned. The criterion for an image $x_i$ to be "clustered" correctly are, it should belong to a cluster which was assigned an identity,and whose identity is same as that of the image $x_i$. Cluster Quality($C_q$) at any point is defined as the ratio of number of images clustered correctly to the total number of images fed for clustering until that point. 
2)Rand Index ($R_{id}$): Rand Index \cite{shortSurvey} is a standard evaluation metric for any clustering algorithm. It is based on the intra-cluster similarity and inter-cluster dissimilarity. For the intra-cluster similarity, if a pair of data vectors is assigned the same cluster in both the target result and the clustering result, then the score will be increased by one. For the inter-cluster dissimilarity, if a pair of vectors is assigned different clusters in both the target result and the clustering result, then the score will be increased by one. On the contrary, if a pair of data vectors is in the same cluster in the target result, but not in the clustering result, the score will not be increased. After we have checked all the possible pairs, the score is normalized by the total number of possible pairs. Exact formulation is given in \cite{shortSurvey}.

The accuracy comparisons are shown  in Fig ~\ref{fig:classClusterQuality}  and Fig ~\ref{fig:classClusterRIndeQuality}. It is evident that our network trained with cluster loss outperforms Trinet in the clustering task.
\label{seqeval}
Choosing the threshold $th$ in the sequential clustering experiment is tricky. Since we are doing a performance evaluation rather than deployment, we tried different thresholds and chose the one that gave best accuracies. This was done for both networks. From the experiments it is observed that our method has better ranking accuracies compared to other existing methods for person ReID, even though by a small margin. Our method excel in a clustering task. This is because the training loss not only tries to bring all similar identities together but also tries to keep the clusters far apart.
\begin{figure}
\begin{tabular}{|p{.4\linewidth}|p{.4\linewidth}|p{.4\linewidth}|p{.4\linewidth}|} 
      \hline
      {\small\centering \hspace{.29\linewidth}Cluster Loss }& {\centering \hspace{.27\linewidth}Triplet Loss}\\
        \hline
      {\includegraphics[width=\linewidth]{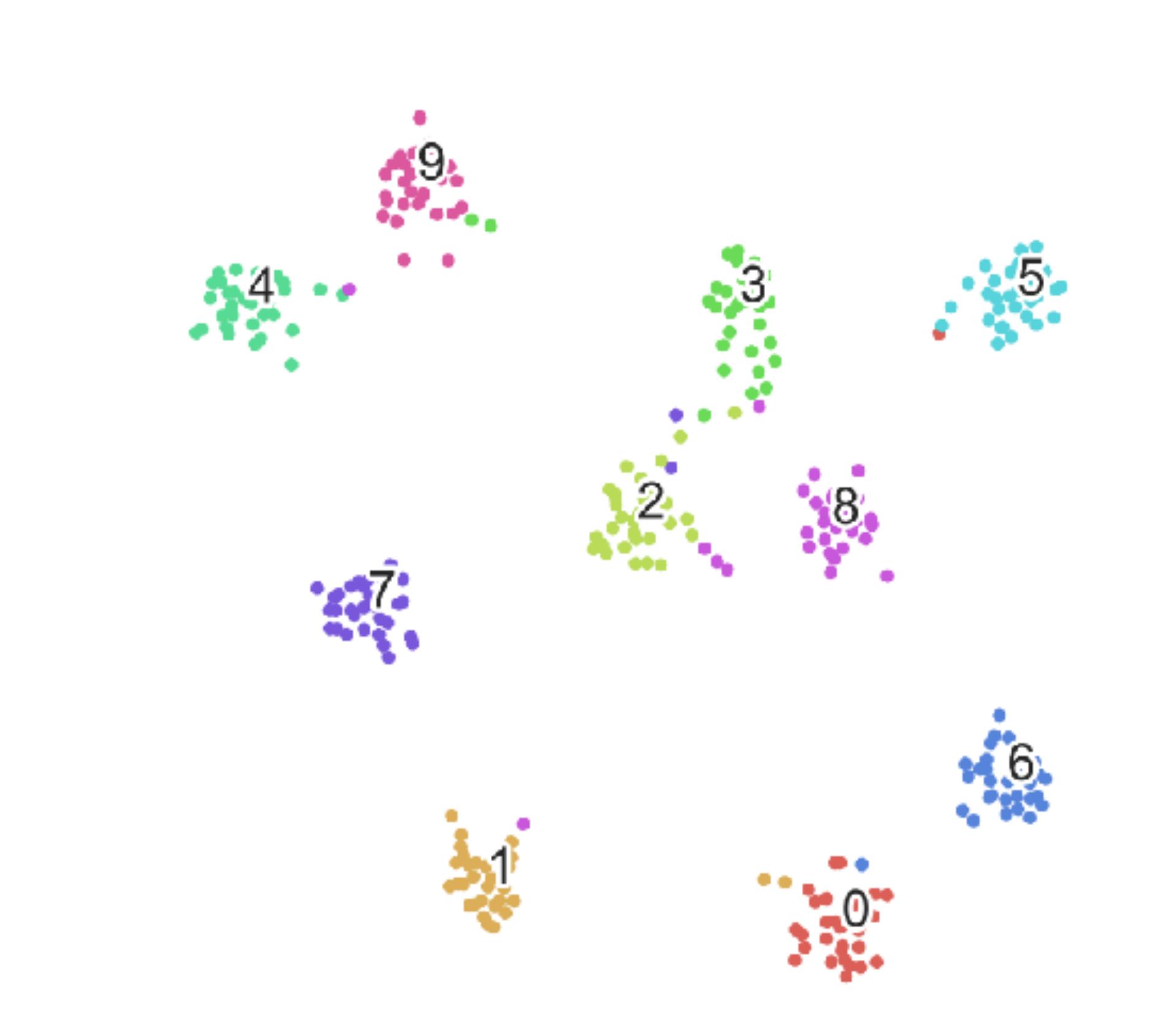}} &
      {\includegraphics[width=\linewidth]{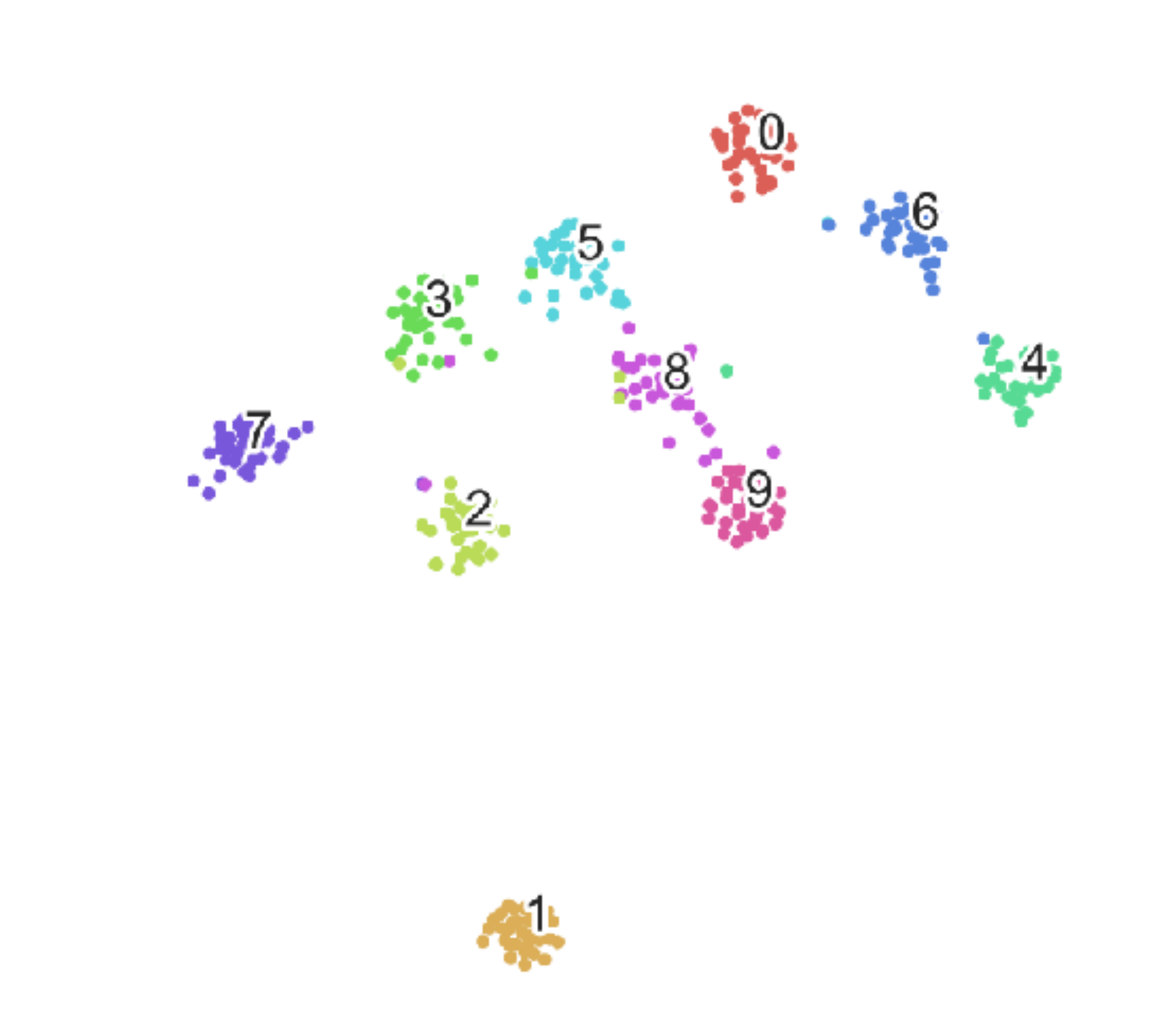}}\\
      {\small\centering \hspace{.47\linewidth}a }& {\small \centering \hspace{.47\linewidth}b}\\
    \hline
      {\includegraphics[width=\linewidth]{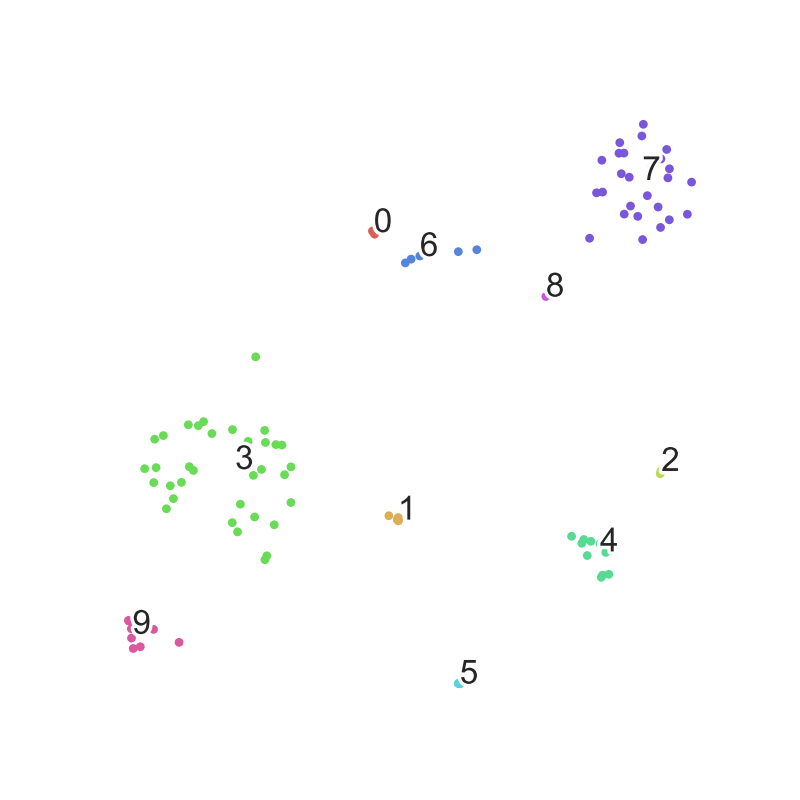}} &
      {\includegraphics[width=\linewidth]{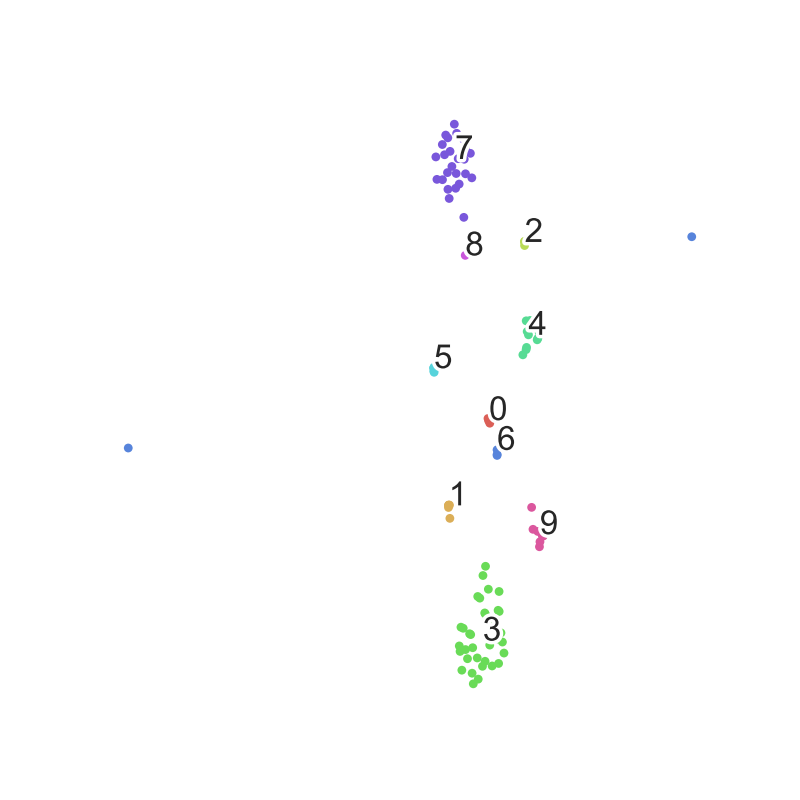}}\\
      {\small\centering \hspace{.47\linewidth} c}& {\small\centering \hspace{.47\linewidth}d}\\
      \hline
      \end{tabular}
 
        \caption{t-SNE plots of the embeddings: (a) and (b) - MNIST valiaation data, (c) and (d) - Randomly sampled 10 identities from Market-1501 test data. }\label{fig:tsnePlot}
\end{figure}

\subsection{Cluster loss - Triplet loss comparison}
We created t-SNE plots for embeddings generated by networks trained using cluster loss and triplet loss to compare the cluster formation. We did this on MNIST\cite{mnist} dataset and Market-1501\cite{market}. For MNIST dataset, a CNN having two convolutional layers with 128 and 256 filters of kernel size 7x7 and 5x5 respectively and embedding dimension of 4, was trained using triplet loss and cluster loss. The t-SNE plot on the validation data is shown in figures \ref{fig:tsnePlot}.a and \ref{fig:tsnePlot}.b. For Market-1501, we used the same networks as used in section \ref{seqeval}. We randomly picked 10 identities from test set and their embeddings are plotted in figures \ref{fig:tsnePlot}.c and \ref{fig:tsnePlot}.d. From the t-SNE plots it is evident that the triplet loss is able to bring the samples from the same class together but the cluster loss performs a better job in separating the clusters.

\section{Conclusion}
In this paper, we introduce cluster loss for training a network for a person ReID task. Training the network with cluster loss shows that it outperforms in learning better parameters for the transformation function which increases the inter class variation and decreases the intra class variation for person re -identification. Our method performs better in ranking tasks compared to the existing state-of-the-art methods. In a clustering task, our network outperformed TriNet~\cite{TriNet} which had the best ranking accuracy among existing methods by a huge margin. In future, as an extension to this work on person ReID, we want to train  and evaluate a cluster loss based network for a face recognition task.

\bibliographystyle{ACM-Reference-Format}
\bibliography{sample-bibliography}


\begin{thebibliography}{00}

%%% the Web address.

\ifx \showCODEN    \undefined \def \showCODEN     #1{\unskip}     \fi
\ifx \showDOI      \undefined \def \showDOI       #1{#1}\fi
\ifx \showISBNx    \undefined \def \showISBNx     #1{\unskip}     \fi
\ifx \showISBNxiii \undefined \def \showISBNxiii  #1{\unskip}     \fi
\ifx \showISSN     \undefined \def \showISSN      #1{\unskip}     \fi
\ifx \showLCCN     \undefined \def \showLCCN      #1{\unskip}     \fi
\ifx \shownote     \undefined \def \shownote      #1{#1}          \fi
\ifx \showarticletitle \undefined \def \showarticletitle #1{#1}   \fi
\ifx \showURL      \undefined \def \showURL       {\relax}        \fi
% invisible to TeX
\providecommand\bibfield[2]{#2}
\providecommand\bibinfo[2]{#2}
\providecommand\natexlab[1]{#1}
\providecommand\showeprint[2][]{arXiv:#2}

\bibitem[\protect\citeauthoryear{Abadi, Barham, Chen, Chen, Davis, Dean, Devin,
  Ghemawat, Irving, Isard, Kudlur, Levenberg, Monga, Moore, Murray, Steiner,
  Tucker, Vasudevan, Warden, Wicke, Yu, and Zheng}{Abadi et~al\mbox{.}}{2016}]%
        {tesnsorflow}
\bibfield{author}{\bibinfo{person}{Mart\'{\i}n Abadi}, \bibinfo{person}{Paul
  Barham}, \bibinfo{person}{Jianmin Chen}, \bibinfo{person}{Zhifeng Chen},
  \bibinfo{person}{Andy Davis}, \bibinfo{person}{Jeffrey Dean},
  \bibinfo{person}{Matthieu Devin}, \bibinfo{person}{Sanjay Ghemawat},
  \bibinfo{person}{Geoffrey Irving}, \bibinfo{person}{Michael Isard},
  \bibinfo{person}{Manjunath Kudlur}, \bibinfo{person}{Josh Levenberg},
  \bibinfo{person}{Rajat Monga}, \bibinfo{person}{Sherry Moore},
  \bibinfo{person}{Derek~G. Murray}, \bibinfo{person}{Benoit Steiner},
  \bibinfo{person}{Paul Tucker}, \bibinfo{person}{Vijay Vasudevan},
  \bibinfo{person}{Pete Warden}, \bibinfo{person}{Martin Wicke},
  \bibinfo{person}{Yuan Yu}, {and} \bibinfo{person}{Xiaoqiang Zheng}.}
  \bibinfo{year}{2016}\natexlab{}.
\newblock \showarticletitle{TensorFlow: A System for Large-scale Machine
  Learning}. In \bibinfo{booktitle}{{\em Proceedings of the 12th USENIX
  Conference on Operating Systems Design and Implementation}} {\em
  (\bibinfo{series}{OSDI'16})}. \bibinfo{publisher}{USENIX Association},
  \bibinfo{address}{Berkeley, CA, USA}, \bibinfo{pages}{265--283}.
\newblock
\showISBNx{978-1-931971-33-1}
\showURL{%
\url{http://dl.acm.org/citation.cfm?id=3026877.3026899}}


\bibitem[\protect\citeauthoryear{Ahmed, Jones, and Marks}{Ahmed
  et~al\mbox{.}}{2015}]%
        {siam2}
\bibfield{author}{\bibinfo{person}{E. Ahmed}, \bibinfo{person}{M. Jones}, {and}
  \bibinfo{person}{T.~K. Marks}.} \bibinfo{year}{2015}\natexlab{}.
\newblock \showarticletitle{An improved deep learning architecture for person
  re-identification}. In \bibinfo{booktitle}{{\em 2015 IEEE Conference on
  Computer Vision and Pattern Recognition (CVPR)}}.
  \bibinfo{pages}{3908--3916}.
\newblock
\showISSN{1063-6919}
\showDOI{%
\url{https://doi.org/10.1109/CVPR.2015.7299016}}


\bibitem[\protect\citeauthoryear{Chen, Yuan, Hua, Zheng, and Wang}{Chen
  et~al\mbox{.}}{2015}]%
        {metric2}
\bibfield{author}{\bibinfo{person}{D. Chen}, \bibinfo{person}{Zejian Yuan},
  \bibinfo{person}{G. Hua}, \bibinfo{person}{N. Zheng}, {and}
  \bibinfo{person}{J. Wang}.} \bibinfo{year}{2015}\natexlab{}.
\newblock \showarticletitle{Similarity learning on an explicit polynomial
  kernel feature map for person re-identification}. In \bibinfo{booktitle}{{\em
  2015 IEEE Conference on Computer Vision and Pattern Recognition (CVPR)}}.
  \bibinfo{pages}{1565--1573}.
\newblock
\showISSN{1063-6919}
\showDOI{%
\url{https://doi.org/10.1109/CVPR.2015.7298764}}


\bibitem[\protect\citeauthoryear{Chen, Guo, and Lai}{Chen
  et~al\mbox{.}}{2016}]%
        {ranking}
\bibfield{author}{\bibinfo{person}{S.~Z. Chen}, \bibinfo{person}{C.~C. Guo},
  {and} \bibinfo{person}{J.~H. Lai}.} \bibinfo{year}{2016}\natexlab{}.
\newblock \showarticletitle{Deep Ranking for Person Re-Identification via Joint
  Representation Learning}.
\newblock \bibinfo{journal}{{\em IEEE Transactions on Image Processing\/}}
  \bibinfo{volume}{25}, \bibinfo{number}{5} (\bibinfo{date}{May}
  \bibinfo{year}{2016}), \bibinfo{pages}{2353--2367}.
\newblock
\showISSN{1057-7149}
\showDOI{%
\url{https://doi.org/10.1109/TIP.2016.2545929}}


\bibitem[\protect\citeauthoryear{Chen, Chen, Zhang, and Huang}{Chen
  et~al\mbox{.}}{2017a}]%
        {quadreplet_loss}
\bibfield{author}{\bibinfo{person}{W. Chen}, \bibinfo{person}{X. Chen},
  \bibinfo{person}{J. Zhang}, {and} \bibinfo{person}{K. Huang}.}
  \bibinfo{year}{2017}\natexlab{a}.
\newblock \showarticletitle{Beyond Triplet Loss: A Deep Quadruplet Network for
  Person Re-identification}. In \bibinfo{booktitle}{{\em 2017 IEEE Conference
  on Computer Vision and Pattern Recognition (CVPR)}}.
  \bibinfo{pages}{1320--1329}.
\newblock
\showISSN{1063-6919}
\showDOI{%
\url{https://doi.org/10.1109/CVPR.2017.145}}


\bibitem[\protect\citeauthoryear{Chen, Chen, Zhang, and Huang}{Chen
  et~al\mbox{.}}{2017b}]%
        {triplet_modi9}
\bibfield{author}{\bibinfo{person}{Weihua Chen}, \bibinfo{person}{Xiaotang
  Chen}, \bibinfo{person}{Jianguo Zhang}, {and} \bibinfo{person}{Kaiqi Huang}.}
  \bibinfo{year}{2017}\natexlab{b}.
\newblock \bibinfo{title}{A Multi-Task Deep Network for Person
  Re-Identification}.
\newblock   (\bibinfo{year}{2017}).
\newblock
\showURL{%
\url{https://aaai.org/ocs/index.php/AAAI/AAAI17/paper/view/14313}}


\bibitem[\protect\citeauthoryear{Cheng, Gong, Zhou, Wang, and Zheng}{Cheng
  et~al\mbox{.}}{2016}]%
        {triplet_modi4}
\bibfield{author}{\bibinfo{person}{D. Cheng}, \bibinfo{person}{Y. Gong},
  \bibinfo{person}{S. Zhou}, \bibinfo{person}{J. Wang}, {and}
  \bibinfo{person}{N. Zheng}.} \bibinfo{year}{2016}\natexlab{}.
\newblock \showarticletitle{Person Re-identification by Multi-Channel
  Parts-Based CNN with Improved Triplet Loss Function}. In
  \bibinfo{booktitle}{{\em 2016 IEEE Conference on Computer Vision and Pattern
  Recognition (CVPR)}}. \bibinfo{pages}{1335--1344}.
\newblock
\showDOI{%
\url{https://doi.org/10.1109/CVPR.2016.149}}


\bibitem[\protect\citeauthoryear{Ding, Lin, Wang, and Chao}{Ding
  et~al\mbox{.}}{2015}]%
        {triplet_modi2}
\bibfield{author}{\bibinfo{person}{Shengyong Ding}, \bibinfo{person}{Liang
  Lin}, \bibinfo{person}{Guangrun Wang}, {and} \bibinfo{person}{Hongyang
  Chao}.} \bibinfo{year}{2015}\natexlab{}.
\newblock \showarticletitle{Deep Feature Learning with Relative Distance
  Comparison for Person Re-identification}.
\newblock \bibinfo{journal}{{\em Pattern Recogn.\/}} \bibinfo{volume}{48},
  \bibinfo{number}{10} (\bibinfo{date}{Oct.} \bibinfo{year}{2015}),
  \bibinfo{pages}{2993--3003}.
\newblock
\showISSN{0031-3203}
\showDOI{%
\url{https://doi.org/10.1016/j.patcog.2015.04.005}}


\bibitem[\protect\citeauthoryear{Farenzena, Bazzani, Perina, Murino, and
  Cristani}{Farenzena et~al\mbox{.}}{2010}]%
        {attr1}
\bibfield{author}{\bibinfo{person}{M. Farenzena}, \bibinfo{person}{L. Bazzani},
  \bibinfo{person}{A. Perina}, \bibinfo{person}{V. Murino}, {and}
  \bibinfo{person}{M. Cristani}.} \bibinfo{year}{2010}\natexlab{}.
\newblock \showarticletitle{Person re-identification by symmetry-driven
  accumulation of local features}. In \bibinfo{booktitle}{{\em 2010 IEEE
  Computer Society Conference on Computer Vision and Pattern Recognition}}.
  \bibinfo{pages}{2360--2367}.
\newblock
\showISSN{1063-6919}
\showDOI{%
\url{https://doi.org/10.1109/CVPR.2010.5539926}}


\bibitem[\protect\citeauthoryear{Geng, Wang, Xiang, and Tian}{Geng
  et~al\mbox{.}}{2016}]%
        {classification_verfic_2}
\bibfield{author}{\bibinfo{person}{Mengyue Geng}, \bibinfo{person}{Yaowei
  Wang}, \bibinfo{person}{Tao Xiang}, {and} \bibinfo{person}{Yonghong Tian}.}
  \bibinfo{year}{2016}\natexlab{}.
\newblock \showarticletitle{Deep Transfer Learning for Person
  Re-identification}.
\newblock \bibinfo{journal}{{\em CoRR\/}}  \bibinfo{volume}{abs/1611.05244}
  (\bibinfo{year}{2016}).
\newblock
\showeprint[arxiv]{1611.05244}
\showURL{%
\url{http://arxiv.org/abs/1611.05244}}


\bibitem[\protect\citeauthoryear{Glorot, Bordes, and Bengio}{Glorot
  et~al\mbox{.}}{2011}]%
        {relu}
\bibfield{author}{\bibinfo{person}{Xavier Glorot}, \bibinfo{person}{Antoine
  Bordes}, {and} \bibinfo{person}{Yoshua Bengio}.}
  \bibinfo{year}{2011}\natexlab{}.
\newblock \showarticletitle{Deep Sparse Rectifier Neural Networks}. In
  \bibinfo{booktitle}{{\em AISTATS}}.
\newblock


\bibitem[\protect\citeauthoryear{Gray and Tao}{Gray and Tao}{2008}]%
        {colorHist1}
\bibfield{author}{\bibinfo{person}{Douglas Gray} {and} \bibinfo{person}{Hai
  Tao}.} \bibinfo{year}{2008}\natexlab{}.
\newblock \showarticletitle{Viewpoint Invariant Pedestrian Recognition with an
  Ensemble of Localized Features}. In \bibinfo{booktitle}{{\em Proceedings of
  the 10th European Conference on Computer Vision: Part I}} {\em
  (\bibinfo{series}{ECCV '08})}. \bibinfo{publisher}{Springer-Verlag},
  \bibinfo{address}{Berlin, Heidelberg}, \bibinfo{pages}{262--275}.
\newblock
\showISBNx{978-3-540-88681-5}
\showDOI{%
\url{https://doi.org/10.1007/978-3-540-88682-2_21}}


\bibitem[\protect\citeauthoryear{Hermans*, Beyer*, and Leibe}{Hermans*
  et~al\mbox{.}}{2017}]%
        {TriNet}
\bibfield{author}{\bibinfo{person}{Alexander Hermans*}, \bibinfo{person}{Lucas
  Beyer*}, {and} \bibinfo{person}{Bastian Leibe}.}
  \bibinfo{year}{2017}\natexlab{}.
\newblock \showarticletitle{{In Defense of the Triplet Loss for Person
  Re-Identification}}.
\newblock \bibinfo{journal}{{\em arXiv preprint arXiv:1703.07737\/}}
  (\bibinfo{year}{2017}).
\newblock


\bibitem[\protect\citeauthoryear{Hirzer, Roth, and Bischof}{Hirzer
  et~al\mbox{.}}{2012a}]%
        {cumMAtcCurve1}
\bibfield{author}{\bibinfo{person}{M. Hirzer}, \bibinfo{person}{P.~M. Roth},
  {and} \bibinfo{person}{H. Bischof}.} \bibinfo{year}{2012}\natexlab{a}.
\newblock \showarticletitle{Person Re-identification by Efficient
  Impostor-Based Metric Learning}. In \bibinfo{booktitle}{{\em 2012 IEEE Ninth
  International Conference on Advanced Video and Signal-Based Surveillance}}.
  \bibinfo{pages}{203--208}.
\newblock
\showDOI{%
\url{https://doi.org/10.1109/AVSS.2012.55}}


\bibitem[\protect\citeauthoryear{Hirzer, Roth, K{\"o}stinger, and
  Bischof}{Hirzer et~al\mbox{.}}{2012b}]%
        {metric1}
\bibfield{author}{\bibinfo{person}{Martin Hirzer}, \bibinfo{person}{Peter~M.
  Roth}, \bibinfo{person}{Martin K{\"o}stinger}, {and} \bibinfo{person}{Horst
  Bischof}.} \bibinfo{year}{2012}\natexlab{b}.
\newblock \showarticletitle{Relaxed Pairwise Learned Metric for Person
  Re-identification}. In \bibinfo{booktitle}{{\em Computer Vision -- ECCV
  2012}}, \bibfield{editor}{\bibinfo{person}{Andrew Fitzgibbon},
  \bibinfo{person}{Svetlana Lazebnik}, \bibinfo{person}{Pietro Perona},
  \bibinfo{person}{Yoichi Sato}, {and} \bibinfo{person}{Cordelia Schmid}}
  (Eds.). \bibinfo{publisher}{Springer Berlin Heidelberg},
  \bibinfo{address}{Berlin, Heidelberg}, \bibinfo{pages}{780--793}.
\newblock
\showISBNx{978-3-642-33783-3}


\bibitem[\protect\citeauthoryear{Ioffe and Szegedy}{Ioffe and Szegedy}{2015}]%
        {batchnorm}
\bibfield{author}{\bibinfo{person}{Sergey Ioffe} {and}
  \bibinfo{person}{Christian Szegedy}.} \bibinfo{year}{2015}\natexlab{}.
\newblock \showarticletitle{Batch Normalization: Accelerating Deep Network
  Training by Reducing Internal Covariate Shift}. In \bibinfo{booktitle}{{\em
  Proceedings of the 32Nd International Conference on International Conference
  on Machine Learning - Volume 37}} {\em (\bibinfo{series}{ICML'15})}.
  \bibinfo{publisher}{JMLR.org}, \bibinfo{pages}{448--456}.
\newblock
\showURL{%
\url{http://dl.acm.org/citation.cfm?id=3045118.3045167}}


\bibitem[\protect\citeauthoryear{Karanam, Gou, Wu, Rates{-}Borras, Camps, and
  Radke}{Karanam et~al\mbox{.}}{2016}]%
        {cumMAtcCurve2}
\bibfield{author}{\bibinfo{person}{Srikrishna Karanam},
  \bibinfo{person}{Mengran Gou}, \bibinfo{person}{Ziyan Wu},
  \bibinfo{person}{Angels Rates{-}Borras}, \bibinfo{person}{Octavia~I. Camps},
  {and} \bibinfo{person}{Richard~J. Radke}.} \bibinfo{year}{2016}\natexlab{}.
\newblock \showarticletitle{A Comprehensive Evaluation and Benchmark for Person
  Re-Identification: Features, Metrics, and Datasets}.
\newblock \bibinfo{journal}{{\em CoRR\/}}  \bibinfo{volume}{abs/1605.09653}
  (\bibinfo{year}{2016}).
\newblock
\showeprint[arxiv]{1605.09653}
\showURL{%
\url{http://arxiv.org/abs/1605.09653}}


\bibitem[\protect\citeauthoryear{Khamis, Kuo, Singh, Shet, and Davis}{Khamis
  et~al\mbox{.}}{2015}]%
        {triplet_modi1}
\bibfield{author}{\bibinfo{person}{Sameh Khamis}, \bibinfo{person}{Cheng-Hao
  Kuo}, \bibinfo{person}{Vivek~K. Singh}, \bibinfo{person}{Vinay~D. Shet},
  {and} \bibinfo{person}{Larry~S. Davis}.} \bibinfo{year}{2015}\natexlab{}.
\newblock \showarticletitle{Joint Learning for Attribute-Consistent Person
  Re-Identification}. In \bibinfo{booktitle}{{\em Computer Vision - ECCV 2014
  Workshops}}, \bibfield{editor}{\bibinfo{person}{Lourdes Agapito},
  \bibinfo{person}{Michael~M. Bronstein}, {and} \bibinfo{person}{Carsten
  Rother}} (Eds.). \bibinfo{publisher}{Springer International Publishing},
  \bibinfo{address}{Cham}, \bibinfo{pages}{134--146}.
\newblock
\showISBNx{978-3-319-16199-0}


\bibitem[\protect\citeauthoryear{Kingma and Ba}{Kingma and Ba}{2014}]%
        {adam}
\bibfield{author}{\bibinfo{person}{Diederik~P. Kingma} {and}
  \bibinfo{person}{Jimmy Ba}.} \bibinfo{year}{2014}\natexlab{}.
\newblock \showarticletitle{Adam: {A} Method for Stochastic Optimization}.
\newblock \bibinfo{journal}{{\em CoRR\/}}  \bibinfo{volume}{abs/1412.6980}
  (\bibinfo{year}{2014}).
\newblock
\showeprint[arxiv]{1412.6980}
\showURL{%
\url{http://arxiv.org/abs/1412.6980}}


\bibitem[\protect\citeauthoryear{Köstinger, Hirzer, Wohlhart, Roth, and
  Bischof}{Köstinger et~al\mbox{.}}{2012}]%
        {kissme}
\bibfield{author}{\bibinfo{person}{M. Köstinger}, \bibinfo{person}{M. Hirzer},
  \bibinfo{person}{P. Wohlhart}, \bibinfo{person}{P.~M. Roth}, {and}
  \bibinfo{person}{H. Bischof}.} \bibinfo{year}{2012}\natexlab{}.
\newblock \showarticletitle{Large scale metric learning from equivalence
  constraints}. In \bibinfo{booktitle}{{\em 2012 IEEE Conference on Computer
  Vision and Pattern Recognition}}. \bibinfo{pages}{2288--2295}.
\newblock
\showISSN{1063-6919}
\showDOI{%
\url{https://doi.org/10.1109/CVPR.2012.6247939}}


\bibitem[\protect\citeauthoryear{LeCun and Cortes}{LeCun and Cortes}{2010}]%
        {mnist}
\bibfield{author}{\bibinfo{person}{Yann LeCun} {and} \bibinfo{person}{Corinna
  Cortes}.} \bibinfo{year}{2010}\natexlab{}.
\newblock \showarticletitle{{MNIST} handwritten digit database}.
\newblock  (\bibinfo{year}{2010}).
\newblock
\showURL{%
\url{http://yann.lecun.com/exdb/mnist/}}


\bibitem[\protect\citeauthoryear{Li, Chen, Zhang, and Huang}{Li
  et~al\mbox{.}}{2017a}]%
        {table3}
\bibfield{author}{\bibinfo{person}{D. Li}, \bibinfo{person}{X. Chen},
  \bibinfo{person}{Z. Zhang}, {and} \bibinfo{person}{K. Huang}.}
  \bibinfo{year}{2017}\natexlab{a}.
\newblock \showarticletitle{Learning Deep Context-Aware Features over Body and
  Latent Parts for Person Re-identification}. In \bibinfo{booktitle}{{\em 2017
  IEEE Conference on Computer Vision and Pattern Recognition (CVPR)}}.
  \bibinfo{pages}{7398--7407}.
\newblock
\showISSN{1063-6919}
\showDOI{%
\url{https://doi.org/10.1109/CVPR.2017.782}}


\bibitem[\protect\citeauthoryear{Li, Zhao, Xiao, and Wang}{Li
  et~al\mbox{.}}{2014}]%
        {siam3}
\bibfield{author}{\bibinfo{person}{W. Li}, \bibinfo{person}{R. Zhao},
  \bibinfo{person}{T. Xiao}, {and} \bibinfo{person}{X. Wang}.}
  \bibinfo{year}{2014}\natexlab{}.
\newblock \showarticletitle{DeepReID: Deep Filter Pairing Neural Network for
  Person Re-identification}. In \bibinfo{booktitle}{{\em 2014 IEEE Conference
  on Computer Vision and Pattern Recognition}}. \bibinfo{pages}{152--159}.
\newblock
\showISSN{1063-6919}
\showDOI{%
\url{https://doi.org/10.1109/CVPR.2014.27}}


\bibitem[\protect\citeauthoryear{Li, Zhu, and Gong}{Li et~al\mbox{.}}{2017b}]%
        {classification_verfic_4}
\bibfield{author}{\bibinfo{person}{Wei Li}, \bibinfo{person}{Xiatian Zhu},
  {and} \bibinfo{person}{Shaogang Gong}.} \bibinfo{year}{2017}\natexlab{b}.
\newblock \showarticletitle{Person Re-Identification by Deep Joint Learning of
  Multi-Loss Classification}.
\newblock \bibinfo{journal}{{\em CoRR\/}}  \bibinfo{volume}{abs/1705.04724}
  (\bibinfo{year}{2017}).
\newblock
\showeprint[arxiv]{1705.04724}
\showURL{%
\url{http://arxiv.org/abs/1705.04724}}


\bibitem[\protect\citeauthoryear{Liao, Hu, Zhu, and Li}{Liao
  et~al\mbox{.}}{2015}]%
        {lomo}
\bibfield{author}{\bibinfo{person}{S. Liao}, \bibinfo{person}{Y. Hu},
  \bibinfo{person}{Xiangyu Zhu}, {and} \bibinfo{person}{S.~Z. Li}.}
  \bibinfo{year}{2015}\natexlab{}.
\newblock \showarticletitle{Person re-identification by Local Maximal
  Occurrence representation and metric learning}. In \bibinfo{booktitle}{{\em
  2015 IEEE Conference on Computer Vision and Pattern Recognition (CVPR)}}.
  \bibinfo{pages}{2197--2206}.
\newblock
\showISSN{1063-6919}
\showDOI{%
\url{https://doi.org/10.1109/CVPR.2015.7298832}}


\bibitem[\protect\citeauthoryear{Lin, Zheng, Zheng, Wu, and Yang}{Lin
  et~al\mbox{.}}{2017}]%
        {table2}
\bibfield{author}{\bibinfo{person}{Yutian Lin}, \bibinfo{person}{Liang Zheng},
  \bibinfo{person}{Zhedong Zheng}, \bibinfo{person}{Yu Wu}, {and}
  \bibinfo{person}{Yi Yang}.} \bibinfo{year}{2017}\natexlab{}.
\newblock \showarticletitle{Improving Person Re-identification by Attribute and
  Identity Learning}.
\newblock \bibinfo{journal}{{\em CoRR\/}}  \bibinfo{volume}{abs/1703.07220}
  (\bibinfo{year}{2017}).
\newblock
\showeprint[arxiv]{1703.07220}
\showURL{%
\url{http://arxiv.org/abs/1703.07220}}


\bibitem[\protect\citeauthoryear{Liu, Feng, Qi, Jiang, and Yan}{Liu
  et~al\mbox{.}}{2017}]%
        {triplet_modi10}
\bibfield{author}{\bibinfo{person}{H. Liu}, \bibinfo{person}{J. Feng},
  \bibinfo{person}{M. Qi}, \bibinfo{person}{J. Jiang}, {and}
  \bibinfo{person}{S. Yan}.} \bibinfo{year}{2017}\natexlab{}.
\newblock \showarticletitle{End-to-End Comparative Attention Networks for
  Person Re-Identification}.
\newblock \bibinfo{journal}{{\em IEEE Transactions on Image Processing\/}}
  \bibinfo{volume}{26}, \bibinfo{number}{7} (\bibinfo{date}{July}
  \bibinfo{year}{2017}), \bibinfo{pages}{3492--3506}.
\newblock
\showISSN{1057-7149}
\showDOI{%
\url{https://doi.org/10.1109/TIP.2017.2700762}}


\bibitem[\protect\citeauthoryear{Liu, Zha, Tian, Liu, Yao, Ling, and Mei}{Liu
  et~al\mbox{.}}{2016}]%
        {triplet_modi8}
\bibfield{author}{\bibinfo{person}{Jiawei Liu}, \bibinfo{person}{Zheng-Jun
  Zha}, \bibinfo{person}{QI Tian}, \bibinfo{person}{Dong Liu},
  \bibinfo{person}{Ting Yao}, \bibinfo{person}{Qiang Ling}, {and}
  \bibinfo{person}{Tao Mei}.} \bibinfo{year}{2016}\natexlab{}.
\newblock \showarticletitle{Multi-Scale Triplet CNN for Person
  Re-Identification}. In \bibinfo{booktitle}{{\em Proceedings of the 2016 ACM
  on Multimedia Conference}} {\em (\bibinfo{series}{MM '16})}.
  \bibinfo{publisher}{ACM}, \bibinfo{address}{New York, NY, USA},
  \bibinfo{pages}{192--196}.
\newblock
\showISBNx{978-1-4503-3603-1}
\showDOI{%
\url{https://doi.org/10.1145/2964284.2967209}}


\bibitem[\protect\citeauthoryear{Liu and Deng}{Liu and Deng}{2015}]%
        {vgg}
\bibfield{author}{\bibinfo{person}{S. Liu} {and} \bibinfo{person}{W. Deng}.}
  \bibinfo{year}{2015}\natexlab{}.
\newblock \showarticletitle{Very deep convolutional neural network based image
  classification using small training sample size}. In \bibinfo{booktitle}{{\em
  2015 3rd IAPR Asian Conference on Pattern Recognition (ACPR)}}.
  \bibinfo{pages}{730--734}.
\newblock
\showISSN{2327-0985}
\showDOI{%
\url{https://doi.org/10.1109/ACPR.2015.7486599}}


\bibitem[\protect\citeauthoryear{Liu, Wang, Wu, Yang, and Yang}{Liu
  et~al\mbox{.}}{2015}]%
        {svm1}
\bibfield{author}{\bibinfo{person}{X. Liu}, \bibinfo{person}{H. Wang},
  \bibinfo{person}{Y. Wu}, \bibinfo{person}{J. Yang}, {and} \bibinfo{person}{M.
  Yang}.} \bibinfo{year}{2015}\natexlab{}.
\newblock \showarticletitle{An Ensemble Color Model for Human
  Re-identification}. In \bibinfo{booktitle}{{\em 2015 IEEE Winter Conference
  on Applications of Computer Vision}}. \bibinfo{pages}{868--875}.
\newblock
\showISSN{1550-5790}
\showDOI{%
\url{https://doi.org/10.1109/WACV.2015.120}}


\bibitem[\protect\citeauthoryear{Ma, Yuen, and Li}{Ma et~al\mbox{.}}{2013}]%
        {colorHist4}
\bibfield{author}{\bibinfo{person}{A.~J. Ma}, \bibinfo{person}{P.~C. Yuen},
  {and} \bibinfo{person}{J. Li}.} \bibinfo{year}{2013}\natexlab{}.
\newblock \showarticletitle{Domain Transfer Support Vector Ranking for Person
  Re-identification without Target Camera Label Information}. In
  \bibinfo{booktitle}{{\em 2013 IEEE International Conference on Computer
  Vision}}. \bibinfo{pages}{3567--3574}.
\newblock
\showISSN{1550-5499}
\showDOI{%
\url{https://doi.org/10.1109/ICCV.2013.443}}


\bibitem[\protect\citeauthoryear{Matsukawa, Okabe, Suzuki, and Sato}{Matsukawa
  et~al\mbox{.}}{2016}]%
        {chuckTable2}
\bibfield{author}{\bibinfo{person}{T. Matsukawa}, \bibinfo{person}{T. Okabe},
  \bibinfo{person}{E. Suzuki}, {and} \bibinfo{person}{Y. Sato}.}
  \bibinfo{year}{2016}\natexlab{}.
\newblock \showarticletitle{Hierarchical Gaussian Descriptor for Person
  Re-identification}. In \bibinfo{booktitle}{{\em 2016 IEEE Conference on
  Computer Vision and Pattern Recognition (CVPR)}}.
  \bibinfo{pages}{1363--1372}.
\newblock
\showISSN{1063-6919}
\showDOI{%
\url{https://doi.org/10.1109/CVPR.2016.152}}


\bibitem[\protect\citeauthoryear{Mignon and Jurie}{Mignon and Jurie}{2012}]%
        {colorHist5}
\bibfield{author}{\bibinfo{person}{A. Mignon} {and} \bibinfo{person}{F.
  Jurie}.} \bibinfo{year}{2012}\natexlab{}.
\newblock \showarticletitle{PCCA: A new approach for distance learning from
  sparse pairwise constraints}. In \bibinfo{booktitle}{{\em 2012 IEEE
  Conference on Computer Vision and Pattern Recognition}}.
  \bibinfo{pages}{2666--2672}.
\newblock
\showISSN{1063-6919}
\showDOI{%
\url{https://doi.org/10.1109/CVPR.2012.6247987}}


\bibitem[\protect\citeauthoryear{Paisitkriangkrai, Shen, and van~den
  Hengel}{Paisitkriangkrai et~al\mbox{.}}{2015}]%
        {triplet_modi3}
\bibfield{author}{\bibinfo{person}{S. Paisitkriangkrai}, \bibinfo{person}{C.
  Shen}, {and} \bibinfo{person}{A. van~den Hengel}.}
  \bibinfo{year}{2015}\natexlab{}.
\newblock \showarticletitle{Learning to rank in person re-identification with
  metric ensembles}. In \bibinfo{booktitle}{{\em 2015 IEEE Conference on
  Computer Vision and Pattern Recognition (CVPR)}}.
  \bibinfo{pages}{1846--1855}.
\newblock
\showISSN{1063-6919}
\showDOI{%
\url{https://doi.org/10.1109/CVPR.2015.7298794}}


\bibitem[\protect\citeauthoryear{Prosser, Zheng, Gong, and Xiang}{Prosser
  et~al\mbox{.}}{2010}]%
        {colorHist2}
\bibfield{author}{\bibinfo{person}{Bryan Prosser}, \bibinfo{person}{Wei-Shi
  Zheng}, \bibinfo{person}{Shaogang Gong}, {and} \bibinfo{person}{Tao Xiang}.}
  \bibinfo{year}{2010}\natexlab{}.
\newblock \showarticletitle{Person Re-Identification by Support Vector
  Ranking}. In \bibinfo{booktitle}{{\em Proceedings of the British Machine
  Vision Conference}}. \bibinfo{publisher}{BMVA Press},
  \bibinfo{pages}{21.1--21.11}.
\newblock
\showISBNx{1-901725-40-5}
\newblock
\shownote{doi:10.5244/C.24.21.}


\bibitem[\protect\citeauthoryear{Schroff, Kalenichenko, and Philbin}{Schroff
  et~al\mbox{.}}{2015}]%
        {facenet}
\bibfield{author}{\bibinfo{person}{F. Schroff}, \bibinfo{person}{D.
  Kalenichenko}, {and} \bibinfo{person}{J. Philbin}.}
  \bibinfo{year}{2015}\natexlab{}.
\newblock \showarticletitle{FaceNet: A unified embedding for face recognition
  and clustering}. In \bibinfo{booktitle}{{\em 2015 IEEE Conference on Computer
  Vision and Pattern Recognition (CVPR)}}. \bibinfo{pages}{815--823}.
\newblock
\showISSN{1063-6919}
\showDOI{%
\url{https://doi.org/10.1109/CVPR.2015.7298682}}


\bibitem[\protect\citeauthoryear{Shi, Yang, Zhu, Liao, Lei, Zheng, and Li}{Shi
  et~al\mbox{.}}{2016}]%
        {triplet_modi6}
\bibfield{author}{\bibinfo{person}{Hailin Shi}, \bibinfo{person}{Yang Yang},
  \bibinfo{person}{Xiangyu Zhu}, \bibinfo{person}{Shengcai Liao},
  \bibinfo{person}{Zhen Lei}, \bibinfo{person}{Wei-Shi Zheng}, {and}
  \bibinfo{person}{Stan~Z. Li}.} \bibinfo{year}{2016}\natexlab{}.
\newblock \showarticletitle{Embedding Deep Metric for Person Re-identification:
  A Study Against Large Variations}. In \bibinfo{booktitle}{{\em ECCV}}.
\newblock


\bibitem[\protect\citeauthoryear{Shi, Hospedales, and Xiang}{Shi
  et~al\mbox{.}}{2015}]%
        {attr2}
\bibfield{author}{\bibinfo{person}{Z. Shi}, \bibinfo{person}{T.~M. Hospedales},
  {and} \bibinfo{person}{T. Xiang}.} \bibinfo{year}{2015}\natexlab{}.
\newblock \showarticletitle{Transferring a semantic representation for person
  re-identification and search}. In \bibinfo{booktitle}{{\em 2015 IEEE
  Conference on Computer Vision and Pattern Recognition (CVPR)}}.
  \bibinfo{pages}{4184--4193}.
\newblock
\showISSN{1063-6919}
\showDOI{%
\url{https://doi.org/10.1109/CVPR.2015.7299046}}


\bibitem[\protect\citeauthoryear{Su, Zhang, Xing, Gao, and Tian}{Su
  et~al\mbox{.}}{2016}]%
        {triplet_modi7}
\bibfield{author}{\bibinfo{person}{Chi Su}, \bibinfo{person}{Shiliang Zhang},
  \bibinfo{person}{Junliang Xing}, \bibinfo{person}{Wen Gao}, {and}
  \bibinfo{person}{Qi Tian}.} \bibinfo{year}{2016}\natexlab{}.
\newblock \showarticletitle{Deep Attributes Driven Multi-camera Person
  Re-identification}. In \bibinfo{booktitle}{{\em ECCV}}.
\newblock


\bibitem[\protect\citeauthoryear{Szegedy, Liu, Jia, Sermanet, Reed, Anguelov,
  Erhan, Vanhoucke, and Rabinovich}{Szegedy et~al\mbox{.}}{2015}]%
        {incpetion}
\bibfield{author}{\bibinfo{person}{C. Szegedy}, \bibinfo{person}{Wei Liu},
  \bibinfo{person}{Yangqing Jia}, \bibinfo{person}{P. Sermanet},
  \bibinfo{person}{S. Reed}, \bibinfo{person}{D. Anguelov}, \bibinfo{person}{D.
  Erhan}, \bibinfo{person}{V. Vanhoucke}, {and} \bibinfo{person}{A.
  Rabinovich}.} \bibinfo{year}{2015}\natexlab{}.
\newblock \showarticletitle{Going deeper with convolutions}. In
  \bibinfo{booktitle}{{\em 2015 IEEE Conference on Computer Vision and Pattern
  Recognition (CVPR)}}. \bibinfo{pages}{1--9}.
\newblock
\showISSN{1063-6919}
\showDOI{%
\url{https://doi.org/10.1109/CVPR.2015.7298594}}


\bibitem[\protect\citeauthoryear{Varior, Haloi, and Wang}{Varior
  et~al\mbox{.}}{2016}]%
        {lstm}
\bibfield{author}{\bibinfo{person}{Rahul~Rama Varior}, \bibinfo{person}{Mrinal
  Haloi}, {and} \bibinfo{person}{Gang Wang}.} \bibinfo{year}{2016}\natexlab{}.
\newblock \showarticletitle{Gated Siamese Convolutional Neural Network
  Architecture for Human Re-identification}. In \bibinfo{booktitle}{{\em
  ECCV}}.
\newblock


\bibitem[\protect\citeauthoryear{Wang, Zuo, Lin, Zhang, and Zhang}{Wang
  et~al\mbox{.}}{2016}]%
        {triplet_modi5}
\bibfield{author}{\bibinfo{person}{F. Wang}, \bibinfo{person}{W. Zuo},
  \bibinfo{person}{L. Lin}, \bibinfo{person}{D. Zhang}, {and}
  \bibinfo{person}{L. Zhang}.} \bibinfo{year}{2016}\natexlab{}.
\newblock \showarticletitle{Joint Learning of Single-Image and Cross-Image
  Representations for Person Re-identification}. In \bibinfo{booktitle}{{\em
  2016 IEEE Conference on Computer Vision and Pattern Recognition (CVPR)}}.
  \bibinfo{pages}{1288--1296}.
\newblock
\showDOI{%
\url{https://doi.org/10.1109/CVPR.2016.144}}


\bibitem[\protect\citeauthoryear{Weinberger and Saul}{Weinberger and
  Saul}{2009}]%
        {tripletMotivation}
\bibfield{author}{\bibinfo{person}{K.~Q. Weinberger} {and}
  \bibinfo{person}{L.~K. Saul}.} \bibinfo{year}{2009}\natexlab{}.
\newblock \showarticletitle{Distance Metric Learning for Large Margin Nearest
  Neighbor Classification}.
\newblock \bibinfo{journal}{{\em JMLR\/}} \bibinfo{volume}{10},
  \bibinfo{number}{1} (\bibinfo{year}{2009}), \bibinfo{pages}{207–244}.
\newblock


\bibitem[\protect\citeauthoryear{Wen, Zhang, Li, and Qiao}{Wen
  et~al\mbox{.}}{2016}]%
        {triplet_bad_intra}
\bibfield{author}{\bibinfo{person}{Yandong Wen}, \bibinfo{person}{Kaipeng
  Zhang}, \bibinfo{person}{Zhifeng Li}, {and} \bibinfo{person}{Yu Qiao}.}
  \bibinfo{year}{2016}\natexlab{}.
\newblock \showarticletitle{A Discriminative Feature Learning Approach for Deep
  Face Recognition}. In \bibinfo{booktitle}{{\em Computer Vision -- ECCV
  2016}}, \bibfield{editor}{\bibinfo{person}{Bastian Leibe},
  \bibinfo{person}{Jiri Matas}, \bibinfo{person}{Nicu Sebe}, {and}
  \bibinfo{person}{Max Welling}} (Eds.). \bibinfo{publisher}{Springer
  International Publishing}, \bibinfo{address}{Cham},
  \bibinfo{pages}{499--515}.
\newblock
\showISBNx{978-3-319-46478-7}


\bibitem[\protect\citeauthoryear{Wong}{Wong}{2015}]%
        {shortSurvey}
\bibfield{author}{\bibinfo{person}{Ka-Chun Wong}.}
  \bibinfo{year}{2015}\natexlab{}.
\newblock \showarticletitle{A short survey on data clustering algorithms}. In
  \bibinfo{booktitle}{{\em Soft Computing and Machine Intelligence (ISCMI),
  2015 Second International Conference on}}. IEEE, \bibinfo{pages}{64--68}.
\newblock


\bibitem[\protect\citeauthoryear{Wu, Shen, and Hengel}{Wu
  et~al\mbox{.}}{2016}]%
        {class1}
\bibfield{author}{\bibinfo{person}{Lin Wu}, \bibinfo{person}{Chunhua Shen},
  {and} \bibinfo{person}{Anton Hengel}.} \bibinfo{year}{2016}\natexlab{}.
\newblock \showarticletitle{PersonNet: Person Re-identification with Deep
  Convolutional Neural Networks}.
\newblock  (\bibinfo{date}{01} \bibinfo{year}{2016}).
\newblock


\bibitem[\protect\citeauthoryear{Wu, Shen, and van~den Hengel}{Wu
  et~al\mbox{.}}{2017}]%
        {deepLDA}
\bibfield{author}{\bibinfo{person}{Lin Wu}, \bibinfo{person}{Chunhua Shen},
  {and} \bibinfo{person}{Anton van~den Hengel}.}
  \bibinfo{year}{2017}\natexlab{}.
\newblock \showarticletitle{Deep linear discriminant analysis on fisher
  networks: A hybrid architecture for person re-identification}.
\newblock \bibinfo{journal}{{\em Pattern Recognition\/}}  \bibinfo{volume}{65}
  (\bibinfo{year}{2017}), \bibinfo{pages}{238 -- 250}.
\newblock
\showISSN{0031-3203}
\showDOI{%
\url{https://doi.org/10.1016/j.patcog.2016.12.022}}


\bibitem[\protect\citeauthoryear{Xiao, Li, Ouyang, and Wang}{Xiao
  et~al\mbox{.}}{2016}]%
        {classficx}
\bibfield{author}{\bibinfo{person}{T. Xiao}, \bibinfo{person}{H. Li},
  \bibinfo{person}{W. Ouyang}, {and} \bibinfo{person}{X. Wang}.}
  \bibinfo{year}{2016}\natexlab{}.
\newblock \showarticletitle{Learning Deep Feature Representations with Domain
  Guided Dropout for Person Re-identification}. In \bibinfo{booktitle}{{\em
  2016 IEEE Conference on Computer Vision and Pattern Recognition (CVPR)}}.
  \bibinfo{pages}{1249--1258}.
\newblock
\showISSN{1063-6919}
\showDOI{%
\url{https://doi.org/10.1109/CVPR.2016.140}}


\bibitem[\protect\citeauthoryear{Xiong, Gou, Camps, and Sznaier}{Xiong
  et~al\mbox{.}}{2014}]%
        {chuckTable}
\bibfield{author}{\bibinfo{person}{Fei Xiong}, \bibinfo{person}{Mengran Gou},
  \bibinfo{person}{Octavia Camps}, {and} \bibinfo{person}{Mario Sznaier}.}
  \bibinfo{year}{2014}\natexlab{}.
\newblock \showarticletitle{Person Re-Identification Using Kernel-Based Metric
  Learning Methods}. In \bibinfo{booktitle}{{\em Computer Vision -- ECCV
  2014}}, \bibfield{editor}{\bibinfo{person}{David Fleet},
  \bibinfo{person}{Tomas Pajdla}, \bibinfo{person}{Bernt Schiele}, {and}
  \bibinfo{person}{Tinne Tuytelaars}} (Eds.). \bibinfo{publisher}{Springer
  International Publishing}, \bibinfo{address}{Cham}, \bibinfo{pages}{1--16}.
\newblock
\showISBNx{978-3-319-10584-0}


\bibitem[\protect\citeauthoryear{Yi, Lei, Liao, and Li}{Yi
  et~al\mbox{.}}{2014}]%
        {siam1}
\bibfield{author}{\bibinfo{person}{Dong Yi}, \bibinfo{person}{Zhen Lei},
  \bibinfo{person}{Shengcai Liao}, {and} \bibinfo{person}{Stan~Z. Li}.}
  \bibinfo{year}{2014}\natexlab{}.
\newblock \showarticletitle{Deep Metric Learning for Person Re-identification}.
  In \bibinfo{booktitle}{{\em Proceedings of the 2014 22Nd International
  Conference on Pattern Recognition}} {\em (\bibinfo{series}{ICPR '14})}.
  \bibinfo{publisher}{IEEE Computer Society}, \bibinfo{address}{Washington, DC,
  USA}, \bibinfo{pages}{34--39}.
\newblock
\showISBNx{978-1-4799-5209-0}
\showDOI{%
\url{https://doi.org/10.1109/ICPR.2014.16}}


\bibitem[\protect\citeauthoryear{Zhang, Xiang, and Gong}{Zhang
  et~al\mbox{.}}{2016b}]%
        {table1}
\bibfield{author}{\bibinfo{person}{L. Zhang}, \bibinfo{person}{T. Xiang}, {and}
  \bibinfo{person}{S. Gong}.} \bibinfo{year}{2016}\natexlab{b}.
\newblock \showarticletitle{Learning a Discriminative Null Space for Person
  Re-identification}. In \bibinfo{booktitle}{{\em 2016 IEEE Conference on
  Computer Vision and Pattern Recognition (CVPR)}}.
  \bibinfo{pages}{1239--1248}.
\newblock
\showISSN{1063-6919}
\showDOI{%
\url{https://doi.org/10.1109/CVPR.2016.139}}


\bibitem[\protect\citeauthoryear{Zhang, Li, Lu, Irie, and Ruan}{Zhang
  et~al\mbox{.}}{2016a}]%
        {svm2}
\bibfield{author}{\bibinfo{person}{Y. Zhang}, \bibinfo{person}{B. Li},
  \bibinfo{person}{H. Lu}, \bibinfo{person}{A. Irie}, {and} \bibinfo{person}{X.
  Ruan}.} \bibinfo{year}{2016}\natexlab{a}.
\newblock \showarticletitle{Sample-Specific SVM Learning for Person
  Re-identification}. In \bibinfo{booktitle}{{\em 2016 IEEE Conference on
  Computer Vision and Pattern Recognition (CVPR)}}.
  \bibinfo{pages}{1278--1287}.
\newblock
\showISSN{1063-6919}
\showDOI{%
\url{https://doi.org/10.1109/CVPR.2016.143}}


\bibitem[\protect\citeauthoryear{Zhao, Ouyang, and Wang}{Zhao
  et~al\mbox{.}}{2014}]%
        {sift}
\bibfield{author}{\bibinfo{person}{R. Zhao}, \bibinfo{person}{W. Ouyang}, {and}
  \bibinfo{person}{X. Wang}.} \bibinfo{year}{2014}\natexlab{}.
\newblock \showarticletitle{Learning Mid-level Filters for Person
  Re-identification}. In \bibinfo{booktitle}{{\em 2014 IEEE Conference on
  Computer Vision and Pattern Recognition}}. \bibinfo{pages}{144--151}.
\newblock
\showISSN{1063-6919}
\showDOI{%
\url{https://doi.org/10.1109/CVPR.2014.26}}


\bibitem[\protect\citeauthoryear{Zheng, Shen, Tian, Wang, Wang, and Tian}{Zheng
  et~al\mbox{.}}{2015}]%
        {market}
\bibfield{author}{\bibinfo{person}{Liang Zheng}, \bibinfo{person}{Liyue Shen},
  \bibinfo{person}{Lu Tian}, \bibinfo{person}{Shengjin Wang},
  \bibinfo{person}{Jingdong Wang}, {and} \bibinfo{person}{Qi Tian}.}
  \bibinfo{year}{2015}\natexlab{}.
\newblock \showarticletitle{Scalable Person Re-identification: A Benchmark}. In
  \bibinfo{booktitle}{{\em Computer Vision, IEEE International Conference on}}.
\newblock


\bibitem[\protect\citeauthoryear{Zheng, Yang, and Hauptmann}{Zheng
  et~al\mbox{.}}{2016a}]%
        {survey1}
\bibfield{author}{\bibinfo{person}{Liang Zheng}, \bibinfo{person}{Yi Yang},
  {and} \bibinfo{person}{Alexander~G. Hauptmann}.}
  \bibinfo{year}{2016}\natexlab{a}.
\newblock \showarticletitle{Person Re-identification: Past, Present and
  Future}.
\newblock \bibinfo{journal}{{\em CoRR\/}}  \bibinfo{volume}{abs/1610.02984}
  (\bibinfo{year}{2016}).
\newblock
\showeprint[arxiv]{1610.02984}
\showURL{%
\url{http://arxiv.org/abs/1610.02984}}


\bibitem[\protect\citeauthoryear{Zheng, Gong, and Xiang}{Zheng
  et~al\mbox{.}}{2013}]%
        {colorHist3}
\bibfield{author}{\bibinfo{person}{W.~S. Zheng}, \bibinfo{person}{S. Gong},
  {and} \bibinfo{person}{T. Xiang}.} \bibinfo{year}{2013}\natexlab{}.
\newblock \showarticletitle{Reidentification by Relative Distance Comparison}.
\newblock \bibinfo{journal}{{\em IEEE Transactions on Pattern Analysis and
  Machine Intelligence\/}} \bibinfo{volume}{35}, \bibinfo{number}{3}
  (\bibinfo{date}{March} \bibinfo{year}{2013}), \bibinfo{pages}{653--668}.
\newblock
\showISSN{0162-8828}
\showDOI{%
\url{https://doi.org/10.1109/TPAMI.2012.138}}


\bibitem[\protect\citeauthoryear{Zheng, Zheng, and Yang}{Zheng
  et~al\mbox{.}}{2016b}]%
        {classification_verfic_3}
\bibfield{author}{\bibinfo{person}{Zhedong Zheng}, \bibinfo{person}{Liang
  Zheng}, {and} \bibinfo{person}{Yi Yang}.} \bibinfo{year}{2016}\natexlab{b}.
\newblock \showarticletitle{A Discriminatively Learned {CNN} Embedding for
  Person Re-identification}.
\newblock \bibinfo{journal}{{\em CoRR\/}}  \bibinfo{volume}{abs/1611.05666}
  (\bibinfo{year}{2016}).
\newblock
\showeprint[arxiv]{1611.05666}
\showURL{%
\url{http://arxiv.org/abs/1611.05666}}


\bibitem[\protect\citeauthoryear{Zhong, Zheng, Cao, and Li}{Zhong
  et~al\mbox{.}}{2017}]%
        {rerank}
\bibfield{author}{\bibinfo{person}{Z. Zhong}, \bibinfo{person}{L. Zheng},
  \bibinfo{person}{D. Cao}, {and} \bibinfo{person}{S. Li}.}
  \bibinfo{year}{2017}\natexlab{}.
\newblock \showarticletitle{Re-ranking Person Re-identification with
  k-Reciprocal Encoding}. In \bibinfo{booktitle}{{\em 2017 IEEE Conference on
  Computer Vision and Pattern Recognition (CVPR)}}.
  \bibinfo{pages}{3652--3661}.
\newblock
\showISSN{1063-6919}
\showDOI{%
\url{https://doi.org/10.1109/CVPR.2017.389}}


\end{thebibliography}

\end{document}